%% file: example_paper.tex
\documentclass{article}

\usepackage{microtype}
\usepackage{graphicx}
\usepackage{subcaption}
\usepackage{booktabs} 

\usepackage{hyperref}

\usepackage[preprint]{icml2026}

\usepackage{amsmath}
\usepackage{amssymb}
\usepackage{mathtools}
\usepackage{amsthm}

\usepackage{xspace}
\usepackage{amsmath}
\usepackage{wrapfig}
\usepackage{caption}     
\usepackage{subcaption}  
\usepackage{enumitem}
\usepackage[table,xcdraw]{xcolor}
\usepackage[breakable]{tcolorbox}
\usepackage{titletoc}
\usepackage{algorithm}
\usepackage{algorithmic}

\usepackage{comment}
\usepackage{multirow}
\usepackage{graphicx}

\usepackage{pifont}
\definecolor{darkgreen}{RGB}{50,100,0}
\definecolor{darkred}{RGB}{200, 0, 0}
\newcommand{\cmark}{\textcolor{darkgreen}{\ding{51}}} %
\newcommand{\xmark}{\textcolor{darkred}{\ding{55}}}

\newcommand{\modelname}{MARS\xspace}
\newcommand{\python}{\raisebox{0pt}{\includegraphics[height=1.0em]{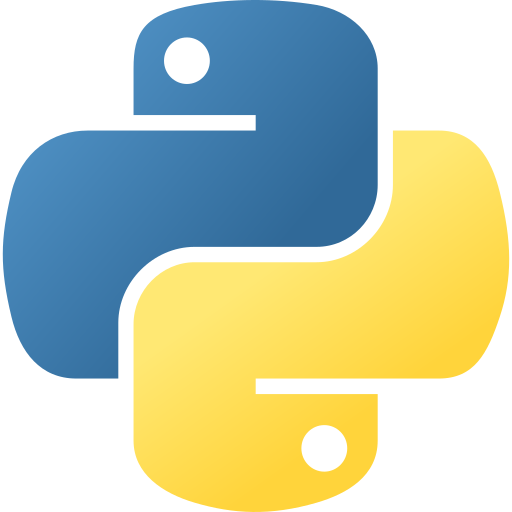}}\xspace}
\newcommand{\google}{\raisebox{0pt}{\includegraphics[height=1.0em]{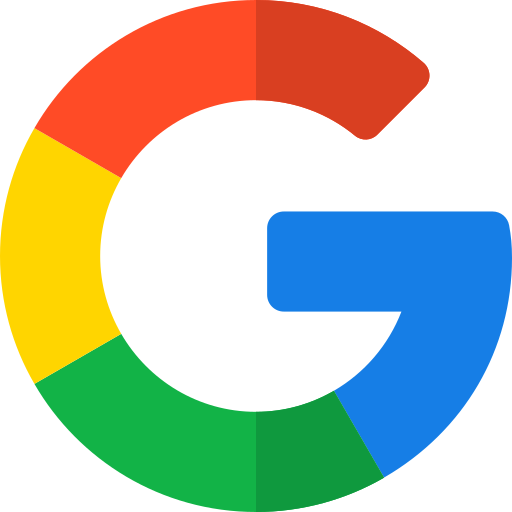}}\xspace}
\newcommand{\scholar}{\raisebox{0pt}{\includegraphics[height=1.0em]{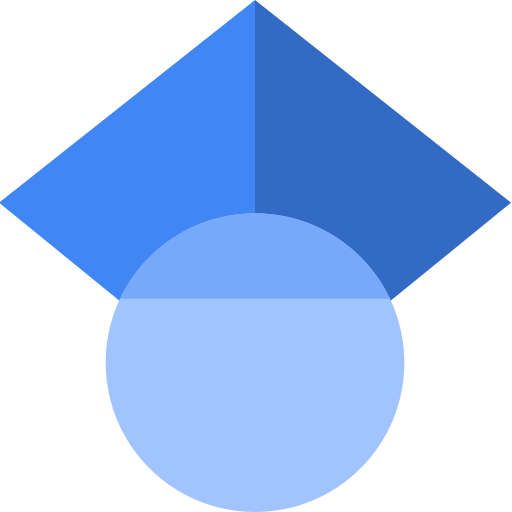}}\xspace}

\usepackage[capitalize,noabbrev]{cleveref}

\theoremstyle{plain}

\theoremstyle{definition}

\theoremstyle{remark}

\usepackage[textsize=tiny]{todonotes}

\icmltitlerunning{MARS: Co-evolving Dual-System Deep Research via Multi-Agent Reinforcement Learning}

\begin{document}

\twocolumn[
  \icmltitle{\modelname: Co-evolving Dual-System Deep Research via\\Multi-Agent Reinforcement Learning}



  \icmlsetsymbol{equal}{*}

  \begin{icmlauthorlist}
    \icmlauthor{Guoxin Chen}{gl,ty}
    \icmlauthor{Zile Qiao}{ty}
    \icmlauthor{Wenqing Wang}{ty}
    \icmlauthor{Donglei Yu}{ty}
    \icmlauthor{Xuanzhong Chen}{ty}
    \icmlauthor{Hao Sun}{ty}
    \icmlauthor{Minpeng Liao}{ty}
    \icmlauthor{Kai Fan}{ty}
    \icmlauthor{Yong Jiang}{ty}
    \icmlauthor{Pengjun Xie}{ty}
    \icmlauthor{Wayne Xin Zhao}{gl}
    \icmlauthor{Ruihua Song}{gl}
    \icmlauthor{Fei Huang}{ty}
  \end{icmlauthorlist}

  \icmlaffiliation{gl}{Gaoling School of Artificial Intelligence, Renmin University of China}
  \icmlaffiliation{ty}{Tongyi Lab, Alibaba Group}

  \icmlcorrespondingauthor{Guoxin Chen}{chenguoxin@ruc.edu.cn}
  \icmlcorrespondingauthor{Xin Zhao}{batmanfly@gmail.com}
  \icmlcorrespondingauthor{Zile Qiao}{qiaozile.qzl@alibaba-inc.com}

  \icmlkeywords{Machine Learning, ICML}

  \vskip 0.3in
]



\printAffiliationsAndNotice{}  

\begin{abstract}
\vspace{-0.1em}
Large Reasoning Models (LRMs) face two fundamental limitations: excessive token consumption when overanalyzing simple information processing tasks, and inability to access up-to-date knowledge beyond their training data.
We introduce \modelname (\textbf{M}ulti-\textbf{A}gent System for Deep \textbf{R}e\textbf{S}earch), a novel \textbf{co-evolution} framework that jointly optimizes dual cognitive systems through multi-agent reinforcement learning.
Unlike prior approaches that employ fixed or independently-trained summarizers, \modelname enables System 1 (fast, intuitive processing) and System 2 (deliberate reasoning) to \textbf{co-adapt} through shared trajectory rewards, developing complementary strategies where System 1 learns to distill information specifically useful for System 2's reasoning.
We extend Group Relative Policy Optimization (GRPO) for multi-agent settings with three key innovations: (1) decoupled gradient computation ensuring proper credit assignment despite shared rewards, (2) bin-packing optimization for efficient parallel information processing, and (3) advantage-weighted balanced sampling preventing training imbalance.
Extensive experiments demonstrate that \modelname (8B), trained under a challenging \textbf{Zero RL} setting without any supervised fine-tuning, achieves 8.17\% on HLE—outperforming WebThinker (32B with SFT, 6.87\%) and narrowing the gap with proprietary models like Claude 3.7 Sonnet (7.89\%)—while achieving an average gain of 8.9\% across 7 knowledge-intensive tasks.
\vspace{-1.7em}
\end{abstract}

\input{section/1_introduction}

\input{section/4_related_work}

\input{section/2_method}

\input{section/3_experiments}

\input{section/5_conclusion}

\section*{Impact Statement}

Our work improves the reliability of deep research by enabling language models to efficiently synthesize up-to-date information from external tools while maintaining strong multi-step reasoning. This capability may benefit knowledge-intensive applications such as education and scientific literature review.
Potential risks include misuse for scalable misinformation or low-quality automated reports, and errors introduced by noisy retrieval or imperfect tool outputs. We therefore emphasize transparent evaluation, careful deployment with human oversight in high-stakes settings, and continued research on robustness and provenance-aware information distillation.

\nocite{langley00}

\bibliography{example_paper}
\bibliographystyle{icml2026}

\newpage
\appendix
\onecolumn

\input{section/6_appendix}



\end{document}

%% file: section/1_introduction.tex
\section{Introduction}\label{sec:intro}

Large Language Models (LLMs) have demonstrated remarkable capabilities across various tasks with their System 1's fast, intuitive thinking, yet they still struggle with complex reasoning tasks~\citep{claude,gpt_4o,yang2024qwen2,gemma,llama4}.
Recent advances in this direction have led to the emergence of Large Reasoning Models (LRMs), which specifically excel at System 2-type, deliberate reasoning when confronted with challenging problems~\citep{jaech2024openai,guo2025deepseek,qwen_qwq,claude_37}.
However, when faced with simpler information processing tasks such as analyzing web pages or summarizing papers, LRMs tend to overanalyze, consuming excessive reasoning tokens that limit their capacity to process large volumes of external information~\citep{chen2024not}.
In contrast, humans effortlessly switch between fast, intuitive processing and slow, deliberate reasoning—a flexibility that current LRMs lack.

Furthermore, developing LLMs capable of complex reasoning in rapidly changing environments remains a significant challenge.
Since model knowledge is confined to the cut-off date of training data, enabling access to external information through retrieval-augmented generation (RAG) is emerging as a promising direction~\citep{c3po2025,li2025search,jin2025search,li2025webthinker,qiao2025webresearcher}.
State-of-the-art proprietary systems~\citep{openai2025deepresearch,google_dr,grok3,kimi-researcher,team2025tongyi} demonstrate the potential of agents that synthesize large volumes of online information for multi-step research tasks.

Recent open-source efforts such as WebThinker~\citep{li2025webthinker} and Search-R1~\citep{jin2025search} have attempted to combine reasoning models with external tools. However, these approaches typically employ \textbf{fixed or independently-trained} information processors that are not optimized for the downstream reasoning task. For example, WebThinker uses a fixed extraction prompt to process retrieved documents, while Search-R1 directly feeds raw search results to the reasoning model. This architectural limitation creates a fundamental mismatch: the information processor cannot learn what specific details the reasoning model needs, leading to either information overload or critical detail loss.

Inspired by the dual-process theory of human cognition~\citep{evans2013dual,frankish2010dual}, we propose \modelname (\textbf{M}ulti-\textbf{A}gent System for Deep \textbf{R}e\textbf{S}earch), a novel framework that enables \textbf{co-evolution} between dual cognitive systems through multi-agent reinforcement learning.
Unlike prior work, \modelname jointly optimizes System 1 (fast, intuitive information processing) and System 2 (deliberate reasoning) through shared trajectory rewards, allowing them to develop complementary strategies.
This co-evolutionary design is motivated by a key insight: \textit{the optimal information distillation strategy depends on the reasoning task, and both systems must co-adapt to achieve effective collaboration}.
Through joint optimization, System 1 learns to extract information that is specifically useful for System 2's reasoning—not generic summaries, but task-relevant insights that emerge from their collaborative training.

The advantages of our co-evolution framework are twofold.
First, through learned specialization, System 1 efficiently filters and distills large volumes of retrieved information (multiple web pages or research papers per query) that would otherwise overwhelm System 2's context capacity, enabling the model to digest comprehensive, up-to-date information without sacrificing reasoning quality.
Second, this dual-system synergy creates an efficient collaborative framework where each component operates within its optimal domain—System 1 handles high-throughput information processing while System 2 focuses on complex multi-step reasoning—resulting in both efficiency and robustness.

To implement our approach, we extend Group Relative Policy Optimization (GRPO)~\citep{shao2024deepseekmath} for multi-agent settings with three key innovations.
First, we design a \textbf{decoupled gradient computation} mechanism where System 1 and System 2 receive gradients from non-overlapping token sets despite sharing trajectory-level rewards, ensuring proper credit assignment.
Second, we employ \textbf{bin-packing optimization} to efficiently organize variable-length retrieved content into optimally-sized chunks, enabling parallel processing by System 1.
Third, we propose \textbf{advantage-weighted balanced sampling} that pre-computes advantages for all samples before balancing, preventing either system from dominating the learning process while preserving statistical integrity.
We also develop a data curation pipeline to collect diverse, graduate-level training examples from public sources, addressing the scarcity of high-quality data for challenging benchmarks like HLE.

Importantly, \modelname operates under a challenging \textbf{``Zero RL''} setting: we train directly from the base model without any Supervised Fine-Tuning (SFT) or policy distillation from stronger models. This contrasts with prior work such as WebThinker (which uses SFT on curated trajectories) and enables direct evaluation of our multi-agent RL framework's effectiveness.
Experiments show that \modelname reaches 8.17\% accuracy on Humanity's Last Exam (HLE) with an 8B model, surpassing WebThinker (32B with SFT, 6.87\%) and approaching Claude 3.7 Sonnet (7.89\%). Across 7 knowledge-intensive QA tasks, it improves performance by 8.9\% on average.
Ablations further show that removing System 1 reduces HLE accuracy by 1.91\%, indicating that System 1 learns task-relevant distillation rather than acting as a passive summarizer.

In summary, our main contributions are:
\begin{itemize}[leftmargin=*, noitemsep, topsep=0pt]
    \item We introduce \modelname, a \textbf{co-evolution} dual-system framework that jointly trains System 1 (information processing) and System 2 (reasoning) with shared trajectory rewards, enabling task-specific information distillation.

    \item We adapt GRPO to multi-agent RL with three techniques: decoupled gradients for credit assignment, bin-packing for parallel information processing, and advantage-weighted balanced sampling for stable joint optimization.

    \item Under a Zero RL setting (no SFT), \modelname (8B) achieves 8.17\% on HLE and improves 7 knowledge-intensive QA tasks; ablations confirm the benefit of co-evolution and each component.
\end{itemize}

%% file: section/4_related_work.tex
\section{Related Work}

\paragraph{Retrieval-Augmented Generation (RAG).}
RAG has emerged as a crucial approach to overcome knowledge limitations of large language models by integrating external information sources~\citep{lewis2020retrieval,nakano2021webgpt,schick2023toolformer,yu2024rankrag,AsaiWWSH24,jiang2024rag,wei2025instructrag,yu2024autorag,c3po2025,li2025webthinker}. However, existing RAG systems face a fundamental dilemma: they either suffer from information overload when processing multiple lengthy documents (such as entire web pages or research papers), or lose critical details when aggressively condensing information~\citep{FanDNWLYCL24,gao2023retrieval,TanD0GFW24}.
Our approach addresses this through a dual-system framework where System 2 handles complex reasoning while System 1 efficiently processes retrieved information, enabling \modelname to manage larger volumes of external information while maintaining reasoning quality.

%% file: section/2_method.tex
\section{Methodology}

This section first overviews the pipeline, highlighting how System 1 and System 2 collaborate for deep research.
We then present our optimization strategies for end-to-end training under a multi-agent reinforcement learning framework.

\subsection{Dual-System Collaborative Framework}

\begin{figure}
    \begin{center}
    \includegraphics[width=\linewidth]{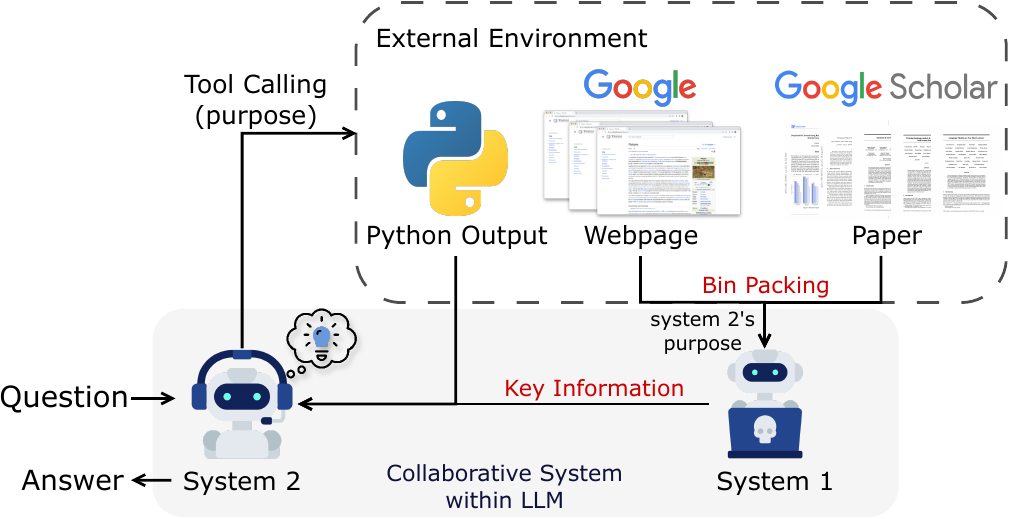}
    \caption{Overview of Dual-System Collaborative Framework.}
    \label{fig:pipeline}
    \end{center}
\end{figure}

We design a collaborative framework that integrates System 1's intuitive processing capabilities with System 2's deliberate reasoning within a unified LLM. As illustrated in Figure~\ref{fig:pipeline}, our framework establishes a synergistic workflow between these two systems to tackle complex questions through external tool utilization.
Specifically, System 2 takes the lead in deliberate reasoning and strategically invokes external tools, while System 1 leverages its intuitive thinking to distill key information from these tool outputs.
The communication between these two systems is facilitated through the ``purpose'' of System 2's current tool invocation. This ``System 2's purpose'' serves as a crucial bridge, allowing System 1 to understand precisely what information to extract and summarize from potentially overwhelming external resources.

\begin{figure*}[t]
    \begin{center}
    \includegraphics[width=\linewidth]{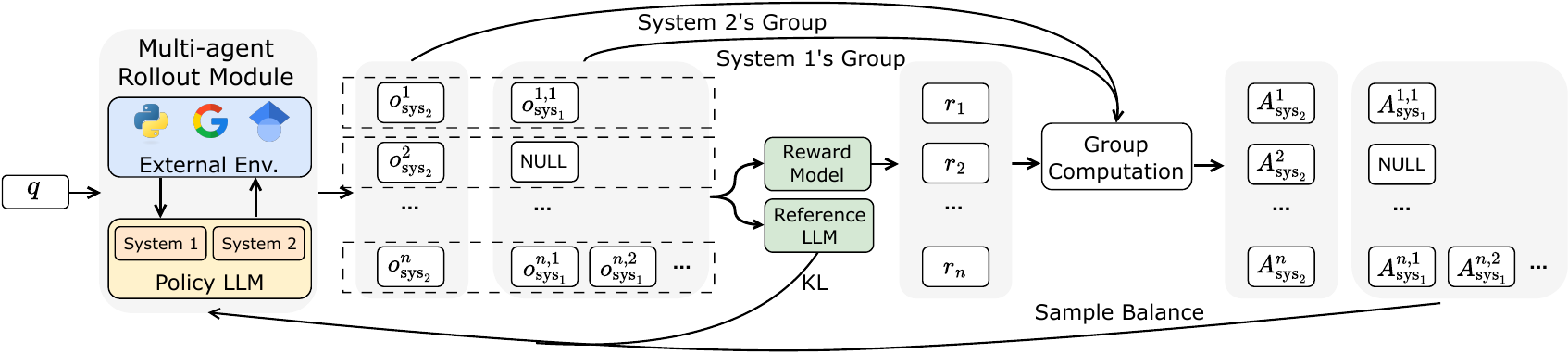}
    \caption{Demonstration of GRPO with multi-agent reinforcement learning in our \modelname.}
    \label{fig:dual_opt}
    \end{center}
\end{figure*}

To formalize this collaborative framework, we represent these two systems activated within the same LLM through different prompts as $\pi_{\text{sys}_1}$ and $\pi_{\text{sys}_2}$.
Given an initial question $q$ as the starting context $c_0$, we model the deep research process as a multi-turn interaction sequence.
While System 2 maintains and reasons with the accumulated context $c_i$ (containing the question and information from previous turns), System 1 operates independently in each turn, focusing solely on processing the current tool outputs without requiring the full historical context.

In the $i$-th turn of interaction, the process unfolds as follows:
(1) System 2 analyzes the current context $c_i$ and generates reasoning steps $s_i$, along with an optional tool request (which includes both the tool parameters $t_i$ and a specific purpose $p_i$):
\begin{equation}
    s_i, (t_i, p_i) = \pi_{\text{sys}_2}(c_i)
\end{equation}
where $t_i$ and $p_i$ can be empty if no tool is needed at this turn.
(2) If $t_i$ is not empty, it is executed by the external environment, producing raw outputs $\{o_{t_i}^{(1)}, o_{t_i}^{(2)}, ..., o_{t_i}^{(n_{t_i})}\}$\footnote{$n_{t_i}$ denotes the number of all outputs from request $t_i$. In our setting, Google Search returns up to 10 web pages per query, Google Scholar returns up to 5 papers, with multiple queries supported in a single tool request.}.
(3) If tool outputs are available, System 1 processes them to extract key information based on System 2's purpose $p_i$. To efficiently handle potentially large volumes of text (e.g., multiple web pages or research papers), we employ bin-packing algorithms~\citep{coffman1984approximation} to organize the variable-length outputs into optimally-sized chunks that can be processed in parallel:
\begin{equation}
    \tilde{o}_{t_i} = \pi_{\text{sys}_1}\left(\text{Bin-Packing}\left(o_{t_i}^{(1)}, o_{t_i}^{(2)}, ..., o_{t_i}^{(n_{t_i})}\right), p_i\right)
\end{equation}
where $\tilde{o}_{t_i}$ represents the distilled information from all tool outputs. The specific bin-packing implementation details are discussed in Section~\ref{sec:bin_packing}.
(4) The context is updated for the next turn by incorporating the reasoning, tool request, purpose, and the distilled information:
\begin{equation}
    c_{i+1} = c_i \oplus \{s_i, t_i, p_i, \tilde{o}_{t_i}\}
\end{equation}
where $\oplus$ denotes context concatenation. Note that for Python Interpreter outputs, $\tilde{o}_{t_i}$ is directly the tool output without System 1 processing, as these results are typically concise and structured.
This process continues iteratively until System 2 determines to answer the original question. The overall generative process can be expressed as:
\begin{equation}
\begin{split}
    \mathcal{P}(&\text{answer}|q) = \prod_{i=1}^{N} \bigg[ \underbrace{\pi_{\text{sys}_2}(s_i, t_i, p_i|c_i)}_{\text{System 2: Reasoning}} \\
    &\cdot \underbrace{\pi_{\text{sys}_1}(\tilde{o}_{t_i}|\text{Bin-Packing}(o_{t_i}^{(1)}, \ldots, o_{t_i}^{(n_{t_i})}), p_i)}_{\text{System 1: Information Processing}} \bigg]
\end{split}
\end{equation}
where $N$ is the total number of turns and the second term is omitted if no tool is called. This formulation clearly expresses how System 2 guides the overall reasoning process, while System 1 efficiently processes external information without maintaining the full context history.

\subsection{Dual-System Optimization Strategies}
To maximize the effectiveness of our dual-system framework, we implement several key optimization strategies:
(1) bin-packing algorithms to enhance System 1's parallel processing efficiency for variable-length retrieved content, and
(2) advantage pre-computation and balanced sampling mechanism, preventing either system from dominating the learning process.
Figure~\ref{fig:dual_opt} illustrates our end-to-end RL training process, from multi-agent rollout to advantage computation and sample balancing.

\subsubsection{Efficient Content Processing with Bin-Packing}\label{sec:bin_packing}
During rollouts, System 2 follows standard token generation for deliberate reasoning and tool-use planning. However, the subsequent tool calls often return multiple outputs of variable length.
Processing these large volumes of variable-length text presents a significant challenge for System 1 due to multiple generations.
To address this, we employ an efficient bin-packing strategy based on the First Fit Decreasing (FFD) algorithm \citep{coffman1984approximation}. 
This approach optimizes the organization of variable-length content into optimally-sized chunks to reduce the number of generations required by System 1.

Specifically, we begin by counting the number of tokens in each tool output $o_{t_i}^{(j)}$.
If the token count exceeds System 1's maximum context length, the output is truncated and placed in an isolated bin. 
For the remaining outputs, we apply the FFD algorithm:
all such outputs are first sorted in decreasing order of their lengths, then assigned to the first bin that can fit them, or to a new bin if no existing one suffices.
We choose FFD over Best Fit Decreasing (BFD) due to its superior efficiency in practice.

\subsubsection{Advantage pre-computation and Balanced Sampling Mechanism}
In our \modelname, following HLE's evaluation prompt~\citep{phan2025humanity}, we employ LLMs as an evaluator to assess the predicted answer of each trajectory.
Note that all System 1 and System 2 samples within the same trajectory share this trajectory-level reward to encourage both systems toward the same goal, rather than pursuing potentially conflicting individual objectives.

For each question $q$, we perform $G$ rollout trajectories and yield exactly $G$ System 2 samples and a variable number of System 1 samples, which depends on the number of tool calls per trajectory and chunks created after Bin-Packing.
This unpredictable imbalance can lead to one system dominating the learning process, potentially undermining the collaborative dynamics essential to our multi-agent system.
To address this, we pre-compute each system's advantage and then balance System 1's samples.
Following GRPO~\citep{shao2024deepseekmath}, rewards are normalized within their corresponding groups to calculate advantage values:
\begin{align}
    A_{\text{sys}_2}^k &= \frac{r_{\text{sys}_2}^k - \text{mean}(\mathbf{r}_{\text{sys}_2})}{\text{std}(\mathbf{r}_{\text{sys}_2})}, &
    A_{\text{sys}_1}^{k,j} &= \frac{r_{\text{sys}_1}^{k,j} - \text{mean}(\mathbf{r}_{\text{sys}_1})}{\text{std}(\mathbf{r}_{\text{sys}_1})},
\end{align}
where $\mathbf{r}_{\text{sys}_2} = \{r_{\text{sys}_2}^1, \ldots, r_{\text{sys}_2}^G\}$ represents all System 2 rewards for the current question, and $\mathbf{r}_{\text{sys}_1} = \{r_{\text{sys}_1}^{k,j} | k \in [1,G], j \in [1,n_k]\}$ represents all System 1 rewards, with $G$ being the group size, $n_k$ being the number of System 1 samples in trajectory $k$, and $r_{\text{sys}_1}^{k,j} = r_{\text{sys}_2}^{k}$.

After computing advantages for all samples, we implement a balanced sampling mechanism to align the number of System 1 samples with System 2. Specifically, if the total number of System 1 samples $M = \sum_{k=1}^{G} n_k$ exceeds $G$, we randomly downsample to exactly $G$ samples; if $M < G$, we upsample through random duplication until reaching $G$ samples.
This pre-computation-then-sampling approach offers two key benefits: First, it ensures that advantage information from all samples contributes to the computation before any sampling occurs, maximizing the utilization of available data. Second, it preserves the statistical integrity of the advantage distribution, as the normalization is performed across the complete set of samples rather than being distorted by the sampling process.

\subsubsection{Multi-Agent Training Objective}

With the balanced samples from both systems, we optimize System 1 and System 2 jointly using an extended GRPO framework, as shown in the Figure~\ref{fig:dual_opt}.
The training samples for each system are distinctly different: 
For System 2, each sample consists of the full reasoning context $c_N = \{s_i, t_i, p_i, \tilde{o}_{t_i}\}_{i=1}^N$, where tokens from System 1's outputs $\tilde{o}_{t_i}$ are masked during loss computation.
For System 1, each sample is a pair of bin-packed input and its corresponding output $(b, \tilde{o})$, where $b$ is a chunk created by the bin-packing algorithm and $\tilde{o}$ is System 1's response.
The overall training objective combines the loss functions for both systems:
\begin{equation}
    \mathcal{L}_{\text{total}} = \mathcal{L}_{\text{sys}_2} + \mathcal{L}_{\text{sys}_1}
\end{equation}
For each system, we apply the GRPO objective~\citep{shao2024deepseekmath}:
\begin{equation}\label{eq:grpo}
    \mathcal{L}_{\text{sys}_i} = \mathbb{E}_{(x,y) \sim \mathcal{D}_i} \left[ \mathcal{L}_{\text{policy}}(x, y, A_{\text{sys}_i}) + \lambda \mathcal{L}_{\text{KL}}(x, y) \right]
\end{equation}
where $\mathcal{D}_i$ represents the balanced dataset for System $i$, $x$ and $y$ denote the input-output pairs, $\mathcal{L}_{\text{policy}}$ is the policy loss and $\mathcal{L}_{\text{KL}}$ is a KL regularization term. 

\textbf{Credit Assignment via Decoupled Gradients.}
A natural question arises: \textit{how can System 1 learn useful summarization behavior when it shares the same trajectory-level reward as System 2?} We address this through \textbf{decoupled gradient computation}, which ensures proper credit assignment despite shared rewards:

\begin{itemize}[leftmargin=*, noitemsep, topsep=0pt]
    \item \textbf{Non-overlapping token sets}: System 2's loss is computed only over reasoning and planning tokens $\{s_i, t_i, p_i\}$, with System 1's output tokens $\tilde{o}_{t_i}$ \textit{masked} during loss computation. System 1's loss is computed only over its distillation outputs, conditioned on the bin-packed input and purpose.

    \item \textbf{Group-specific advantage normalization}: Each system's advantage (Eq. 5-6) is normalized within its own sample group. This means System 1 learns ``what made my summary better than other summaries in this group?'' rather than simply ``was the overall trajectory successful?''
\end{itemize}

Together, the shared reward aligns both systems toward correct problem solving, while decoupled gradients provide role-specific learning signals: System 2 improves planning and reasoning, and System 1 improves the utility of its distillation.
This enables co-evolution: System 2 learns better purposes $p_i$, and System 1 learns more task-relevant extractions.

%% file: section/3_experiments.tex
\section{Experiments}\label{sec:exp}

\subsection{Evaluation Datasets and Metrics}
We evaluate our \modelname on several challenging benchmarks that require sophisticated multi-step reasoning and external knowledge:
(1) \textbf{Humanity’s Last Exam (HLE)}~\citep{phan2025humanity} is an extremely challenging dataset containing advanced problems. We utilize its text-only subset with 2,154 questions.
(2) \textbf{Single-Hop Question Answering}, including NQ~\citep{KwiatkowskiPRCP19}, TriviaQA~\citep{JoshiCWZ17}, and PopQA~\citep{MallenAZDKH23}.
(3) \textbf{Multi-Hop Question Answering}, including HotpotQA~\citep{yang-etal-2018-hotpotqa}, 2WikiMultiHopQA~\citep{HoNSA20}, Musique~\citep{trivedi-etal-2022-musique}, and Bamboogle~\citep{PressZMSSL23}. 
For evaluation, we follow different protocols for different benchmarks. For HLE, we adopt its official evaluation prompt~\citep{phan2025humanity} with GPT-4o~\citep{gpt_4o} as the judge. For other QA datasets, following~\cite{c3po2025}, we employ Qwen2.5-72B-Instruct~\citep{yang2024qwen2} as the evaluation model.

\begin{table}[t]
\centering
\caption{Comparison of training configurations across methods. \modelname achieves superior results despite using smaller models and no SFT warmup.}
\label{tab:baseline_config}
\small
\begin{tabular}{@{}lcccc@{}}
\toprule
\textbf{Method} & \textbf{Model Size} & \textbf{SFT} & \textbf{RL} & \textbf{Co-evolution} \\
\midrule
WebThinker & 32B & \cmark & \cmark & \xmark \\
C-3PO & 72B & \cmark & \cmark & \xmark \\
Search-R1 & 7B & \xmark & \cmark & \xmark \\
Search-o1 & 7B & \xmark & \xmark & \xmark \\
\midrule
\modelname (Ours) & 7B/8B & \xmark & \cmark & \cmark \\
\bottomrule
\end{tabular}
\end{table}

\input{tables/hle}

\input{tables/mhqa}
\input{figures/rl_process}

\subsection{Baselines and Implementation Details}\label{sec:impl}
To evaluate the effectiveness of \modelname, we compare our method with the following baselines.
(1) \textbf{Direct Reasoning}: Models that directly answer questions without external knowledge, including open-source models (Qwen2.5 series~\citep{yang2024qwen2} and QwQ-32B~\citep{qwen_qwq}) and powerful proprietary models (DeepSeek-R1-671B~\citep{guo2025deepseek}, GPT-4o~\citep{gpt_4o}, o1~\citep{2409_openai_o1}, Claude 3.7 Sonnet~\citep{claude_37}, Gemini 2.5 Pro~\citep{gemini_25_pro}, o3(high)~\citep{openai_o3_mini}), and o4-mini(high)~\citep{openai_o3_o4_mini}.
(2) \textbf{Advanced RAG Reasoning Methods}: We consider the standard RAG that retrieve top-10 documents based on the question, and several iterative RAG methods, including Self-RAG~\citep{AsaiWWSH24}, InstructRAG~\citep{wei2025instructrag}, Auto-RAG~\citep{yu2024autorag} and C-3PO~\citep{c3po2025}.
(3) \textbf{R1-like Reasoning with Search}: Methods that integrate external knowledge into R1-like reasoning, including Search-o1~\citep{li2025search}, Search-R1~\citep{jin2025search}, WebThinker~\citep{li2025webthinker}, and OpenAI Deep Research~\citep{openai2025deepresearch}.

\textbf{Implementation and Fair Comparison.}
We initialize our policy model with Qwen2.5-7B-Instruct~\citep{yang2024qwen2} and Qwen3-8B~\citep{yang2025qwen3} and design dedicated prompts for System 1 and System 2 (Appendix~\ref{app:prompt}) to facilitate multi-agent reinforcement learning.
Crucially, \modelname operates under a \textbf{``Zero RL''} setting: we train directly from the base model \textit{without any Supervised Fine-Tuning (SFT)} or policy distillation from stronger models. This contrasts with baselines such as WebThinker, which employs SFT on curated trajectories before RL training.
Table~\ref{tab:baseline_config} summarizes the key differences in training configurations. Despite using smaller models (7B/8B vs. 32B/72B) and no SFT warmup, \modelname achieves superior performance, validating the effectiveness of our multi-agent RL framework rather than relying on model scale or supervised data.
All training data are curated from public sources (Appendix~\ref{app:data_filter}), ensuring data source parity with baselines. Additional implementation details are provided in Appendix~\ref{app:imple_details}.

\subsection{Main Results on Humanity's Last Exam}
Table~\ref{tab:main_results_hle} compares \modelname with various baselines on HLE, a challenging benchmark requiring sophisticated reasoning and up-to-date knowledge across multiple disciplines.

\textbf{\modelname achieves 7.38\% accuracy (8.17\% with Qwen3-8B), a 3.86\% improvement over the base model}, outperforming larger models including WebThinker (32B) and C-3PO (72B). The performance gap with proprietary models like Claude 3.7 Sonnet (7.89\%) and o1 (7.75\%) is notably small, especially considering \modelname uses only 7B/8B parameters.
Beyond the overall score, Table~\ref{tab:main_results_hle} suggests that our gains are broad rather than isolated to a single subject area, consistent with the goal of building a general deep-research agent. We attribute these improvements to the complementary roles of the two systems: System 2 concentrates its long-form reasoning budget on planning and synthesis, while System 1 absorbs the high-throughput reading burden by filtering and compressing retrieved content.
Importantly, this separation mitigates a common failure mode of tool-augmented reasoning---context dilution. Instead of appending raw documents into the reasoning context, System 1 produces purpose-conditioned evidence snippets, which makes subsequent reasoning steps more stable and less sensitive to retrieval noise.

\subsection{Main Results on Knowledge-intensive Reasoning}
Table~\ref{tab:main_results_mhqa} presents results on 7 knowledge-intensive QA tasks covering both single-hop and multi-hop reasoning.
\modelname consistently outperforms all baselines, achieving \textbf{8.95\% average improvement over C-3PO}. The gains are particularly pronounced on multi-hop tasks (+12.2\% over C-3PO), demonstrating that our framework excels at complex reasoning chains that require multiple retrieval--reasoning iterations.
These results highlight that the benefit of co-evolution is not limited to HLE. On single-hop QA, our method improves factual recall by using lightweight, purpose-guided retrieval and distillation, reducing hallucinations caused by relying on stale parametric knowledge. On multi-hop QA, the advantage is larger because errors compound across steps: missing one entity or relation early can derail the entire chain.

We attribute the improvements to two complementary behaviors learned during joint training. First, System 2 learns to decompose questions into targeted sub-queries and to interleave tool use with reasoning, maintaining a coherent reasoning trace instead of attempting to ``read everything'' in-context. Second, System 1 learns purpose-conditioned distillation that preserves entities, relations, and intermediate constraints (rather than producing generic summaries), which is critical for multi-hop settings where fine-grained details determine whether subsequent hops are correct.
Overall, these trends suggest that \modelname improves both \emph{evidence acquisition} (better tool-use decisions) and \emph{evidence utilization} (more faithful, task-relevant distillation), leading to robust gains across diverse knowledge-intensive benchmarks.

\subsection{Analysis of Multi-agent RL Process}
Figure~\ref{fig:rl_process} illustrates key aspects of our multi-agent RL training dynamics.
We observe consistent improvement in HLE score from approximately 2\% to over 10\%, correlating with reward stabilization around 0.4. Meanwhile, the average number of tools used per question increases from $\sim$1 to over 2, suggesting a shift from ``single-shot'' retrieval toward iterative evidence gathering.

\input{tables/ablation_tool}
\begin{table*}[t]
\centering
\caption{Ablation on System 1 co-evolution (HLE).}
\label{tab:ablation_sys1}
\small
\begin{tabular}{@{}lccccccccc@{}}
\toprule
\textbf{Method} & \textbf{Bio/Med} & \textbf{Chem.} & \textbf{CS/AI} & \textbf{Eng.} & \textbf{Hum.} & \textbf{Math} & \textbf{Phys.} & \textbf{Other} & \textbf{Avg.} \\
\midrule
Qwen2.5-7B (Base) & 5.42 & 3.00 & 1.76 & 3.22 & 4.66 & 3.58 & 1.98 & 4.00 & 3.52 \\
\modelname (Full) & \underline{\textbf{12.66}} & 3.00 & \underline{\textbf{5.75}} & 4.83 & \underline{\textbf{11.92}} & \underline{\textbf{6.46}} & \underline{\textbf{6.43}} & \underline{\textbf{7.42}} & \underline{\textbf{7.38}} \\
\quad w/o System 1 & 9.95 & \underline{\textbf{5.00}} & 3.98 & \underline{\textbf{6.45}} & 7.77 & 4.92 & 3.46 & 4.57 & 5.47 \\
\bottomrule
\end{tabular}
\end{table*}

\input{tables/pages_per_ques}

Beyond the overall upward trend, Figure~\ref{fig:rl_process} reveals two qualitative phases. In early training, the model is conservative in tool use and often under-explores external evidence, leading to low reward and slow score improvement. As training proceeds, the policy learns to invoke tools more frequently on harder questions and to sustain longer interaction trajectories; this coincides with higher reward and improved accuracy.
Notably, these trends are aligned: as training progresses, the model not only becomes better at deciding \emph{when} to use tools, but also improves \emph{how} to use them (i.e., decomposing a question into multiple targeted queries and integrating the resulting evidence). This is consistent with our design goal that System 2 should specialize in planning/tool-use, while System 1 specializes in high-throughput information processing.
We find this behavior important because naive tool-augmented reasoning often fails due to noisy retrieval: issuing more tool calls can easily increase irrelevant context and distract reasoning. In \modelname, System 1 mitigates this risk by purpose-conditioned distillation, while the balanced sampling and decoupled-gradient design ensures that System 1 is directly optimized for \emph{utility} to System 2 rather than for generic summarization.
Extended analysis of RL dynamics (tool preferences and response-length trends) is provided in Appendix~\ref{app:rl_extended}.

\subsection{Ablation Study}

\textbf{Tool Ablation on HLE.}
Table~\ref{tab:ablation_tool} shows that using all three tools achieves the best performance (7.38\%), highlighting that complex questions often require \emph{both} reliable retrieval (Search/Scholar) and precise computation (Python).
Several patterns are worth noting. (1) \textbf{Google Search is the strongest single tool}: removing Search yields the largest overall drop, consistent with Search providing broad, high-recall coverage across domains. (2) \textbf{Python is domain-specialized}: its removal hurts Math/Physics most, indicating that many questions require exact arithmetic, symbolic manipulation, or unit-aware calculations that are brittle under purely textual reasoning. (3) \textbf{Scholar complements Search}: although Scholar is used less frequently, it benefits research-heavy categories (e.g., CS/AI, Other) where reliable answers may depend on technical terminology and up-to-date academic evidence.
A key takeaway is that tools are not simply additive: removing a tool can sometimes improve a subset of categories, reflecting that the model must learn both \emph{tool selection} and \emph{tool orchestration}. This can happen when a weaker tool policy introduces noisy evidence or distracts the reasoning process. Our joint training setup mitigates this by encouraging System 2 to learn \emph{when} a tool is worth the latency/cost and by training System 1 to extract only task-relevant evidence from tool outputs, reducing the chance that retrieval noise dilutes System 2's context.

\textbf{Information Volume.}
Table~\ref{tab:pages_per_ques} shows that \modelname processes substantial external information per question (e.g., many full web pages/papers), motivating the need for System 1's efficient distillation.
This analysis highlights a practical constraint for deep research: naively concatenating retrieved documents quickly saturates the context window and degrades reasoning due to irrelevant or redundant text. Our bin-packing + System 1 distillation pipeline makes the information volume tractable by converting many long documents into a small set of evidence-focused chunks that System 2 can reliably consume.

\subsection{Ablation on System 1 Co-evolution}\label{sec:ablation_sys1}

To validate whether System 1 actively contributes through co-evolution (vs. passive compression), we conduct an ablation where System 2 processes raw tool outputs directly without System 1's distillation.
As shown in Table~\ref{tab:ablation_sys1}, removing System 1 causes a \textbf{1.91\% performance drop} (7.38\%$\to$5.47\%). This confirms that System 1 \textbf{actively co-evolves} with System 2, learning to extract task-relevant information that specifically supports reasoning—beyond what fixed extraction prompts could achieve.
Importantly, this ablation maintains access to all tools but removes learned collaboration, confirming that our \textbf{MARL-based co-evolution}—not merely tool availability—is the key driver of improvement.


%% file: tables/hle.tex
\begin{table*}[t]
\centering
\caption{Main Results on HLE (evaluated with official evaluation prompt~\citep{phan2025humanity} by GPT-4o). Results for proprietary models are from the official leaderboard for reference. For open-source models, the best results are in \textbf{bold} and the second-best are \underline{underlined}. \modelname operates under Zero RL (no SFT) with smaller models yet outperforms larger SFT-trained baselines.}
\small
\label{tab:main_results_hle} 
\resizebox{0.95\linewidth}{!}{
\begin{tabular}{@{}lccccccccc@{}}
\toprule
                                  & \multicolumn{9}{c}{\textbf{Humanity's Last Exam}}                                                                                                                                \\ \cmidrule(l){2-10} 
\multirow{-2}{*}{\textbf{Method}} & \textbf{Bio/Med}  & \textbf{Chem.}   & \textbf{CS/AI}   & \textbf{Engineering} & \textbf{Humanities} & \textbf{Math}    & \textbf{Physics} & \textbf{Other}   & \textbf{Avg.}    \\ \midrule
\multicolumn{10}{c}{\textit{\textbf{Proprietary Models (For Reference)}}}                                                                                                                                            \\ \midrule
OpenAI Deep Research              & -                 & -                & -                & -                    & -                   & -                & -                & -                & 26.60            \\
o3 (high)                         & -                 & -                & -                & -                    & -                   & -                & -                & -                & 20.57            \\
o4-mini (high)                    & -                 & -                & -                & -                    & -                   & -                & -                & -                & 18.90            \\
Gemini 2.5 Pro                    & -                 & -                & -                & -                    & -                   & -                & -                & -                & 18.38            \\
Deepseek R1                       & -                 & -                & -                & -                    & -                   & -                & -                & -                & 8.54             \\
Claude 3.7 Sonnet                 & -                 & -                & -                & -                    & -                   & -                & -                & -                & 7.89             \\
o1                                & -                 & -                & -                & -                    & -                   & -                & -                & -                & 7.75             \\
GPT-4.1                           & -                 & -                & -                & -                    & -                   & -                & -                & -                & 4.91             \\
GPT-4o                            & -                 & -                & -                & -                    & -                   & -                & -                & -                & 2.32             \\ \midrule
\multicolumn{10}{c}{\textit{\textbf{Open-Source Models}}}                                                                                                                                                            \\ \midrule
QwQ-32B                           & 9.05              & 6.00             & 4.86 & 1.61                 & 6.21                & 4.92             & 4.95             & 3.42             & 5.28             \\
Qwen2.5-72B                       & 11.31             & 6.00             & 1.76             & 1.61                 & 7.25                & 3.07             & 3.96             & 2.28             & 4.27             \\
Qwen2.5-7B                        & 5.42              & 3.00             & 1.76             & 3.22                 & 4.66                & 3.58             & 1.98             & 4.00             & 3.52             \\
Qwen3-8B                          & 6.78              & 4.00             & 3.53             & 3.22                 & 5.69                & 4.21             & 3.46             & 2.85             & 4.31             \\ \midrule
\multicolumn{10}{c}{\textit{\textbf{R1-like Reasoning with Search}}}                                                                                                                                                 \\ \midrule
WebThinker(QwQ-32B)               & \textbf{14.47}             & \textbf{8.00}    & 4.42             & \underline{6.45}     & \underline{10.88}               & 4.51             & 1.98             & \textbf{14.28}   & 6.87             \\
C-3PO (Qwen2.5-72B)               & 9.95              & \underline{7.00} & 4.86 & \textbf{9.67}        & 4.66                & 5.43             & 3.46             & 5.71             & 5.79             \\
Search-o1(Qwen2.5-7B)             & 9.95              & 2.00             & 2.67             & 1.61                 & 4.14                & 4.41             & 3.98             & 7.42             & 4.79             \\
Search-R1(Qwen2.5-7B)             & 6.33              & 6.00             & 4.42             & \underline{6.45}     & 3.62                & 3.59             & 1.48             & 4.00             & 3.99             \\ \midrule
\rowcolor[HTML]{D8ECE4} 
\modelname (Qwen2.5-7B)           & 12.66 & 3.00             & \underline{5.75}             & 4.83                 & \textbf{11.92}   & \underline{6.46} & \underline{6.43} & 7.42             & \underline{7.38} \\
\rowcolor[HTML]{D8ECE4} 
\modelname (Qwen3-8B)             & \underline{13.12}    & 6.00             & \textbf{8.84}    & \underline{6.45}     & 8.29       & \textbf{7.17}    & \textbf{7.92}    & \underline{8.57} & \textbf{8.17}    \\ \bottomrule
\end{tabular}
}
\end{table*}

%% file: tables/mhqa.tex
\begin{table*}
\centering
\scriptsize
\caption{Main Results on Knowledge-intensive Tasks.}
\small
\label{tab:main_results_mhqa} 
\resizebox{0.95\linewidth}{!}{
\begin{tabular}{@{}lcccccccc@{}}
\toprule
                                  & \multicolumn{3}{c}{\textbf{Single-Hop QA}}              & \multicolumn{4}{c}{\textbf{Multi-Hop QA}}                                    &                                 \\ \cmidrule(lr){2-8}
\multirow{-2}{*}{\textbf{Method}} & \textbf{NQ}      & \textbf{TriviaQA} & \textbf{PopQA}   & \textbf{HotpotQA} & \textbf{2Wiki}   & \textbf{Musique} & \textbf{Bamboogle} & \multirow{-2}{*}{\textbf{Avg.}} \\ \midrule
Direct Answer                     & 29.6             & 50.0              & 30.8             & 30.6              & 28.4             & 13.4             & 30.4               & 30.45                           \\
Standard RAG                      & 45.6             & 72.0              & 46.2             & 43.4              & 29.4             & 25.6             & 41.6               & 43.40                           \\ \midrule
Self-RAG                          & 49.4             & 66.2              & 38.6             & -                 & -                & -                & -                  & 51.40                           \\
InstructRAG                       & 47.8             & 66.6              & 39.6             & -                 & 35.8             & -                & -                  & 47.45                           \\
Auto-RAG                          & 52.4             & 62.2              & 36.8             & 44.4              & 46.8             & -                & -                  & 48.52                           \\ \midrule
Search-o1                         & 42.4             & 64.6              & 38.2             & 46.8              & 52.8             & 26.2             & 56.0               & 46.71                           \\
Search-R1                         & 51.6             & 72.6              & 56.8             & 46.6              & 50.4             & 28.4             & 57.4               & 51.97                           \\
C-3PO                             & 47.4             & 75.0              & 56.8             & 48.2              & 52.6             & 33.2             & 60.8               & 53.42                           \\ \midrule
\rowcolor[HTML]{D8ECE4} 
MARS (Qwen2.5-7B)                 & \underline{60.6} & \underline{76.4}  & \underline{64.4} & \underline{60.4}  & \underline{66.4} & \underline{39.6} & \underline{68.8}   & \underline{62.37}               \\
\rowcolor[HTML]{D8ECE4} 
MARS (Qwen3-8B)                   & \textbf{63.2}    & \textbf{78.6}     & \textbf{66.8}    & \textbf{63.6}     & \textbf{69.2}    & \textbf{42.4}    & \textbf{71.2}      & \textbf{65.00}                  \\ \bottomrule
\end{tabular}
}
\end{table*}

%% file: figures/rl_process.tex
\begin{figure*}[t]
  \centering
  \begin{subfigure}[b]{0.32\textwidth}
    \includegraphics[width=\textwidth]{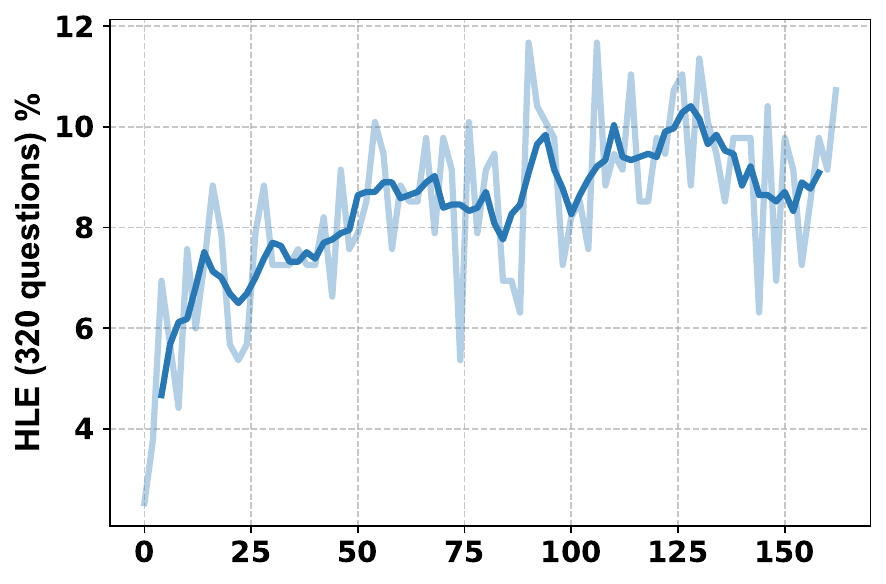}
    \caption{HLE Score Progression}
    \label{fig:hle_score}
  \end{subfigure}
  \hfill
  \begin{subfigure}[b]{0.32\textwidth}
    \includegraphics[width=\textwidth]{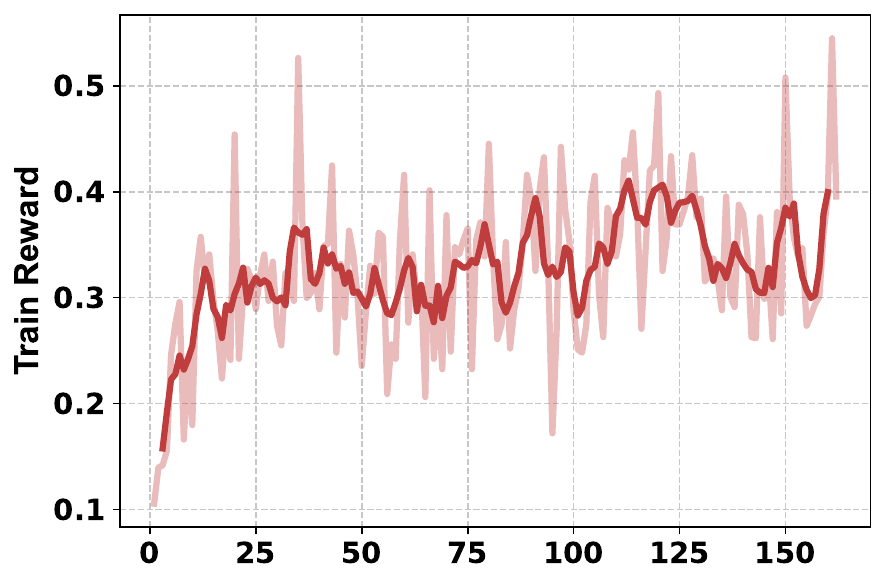}
    \caption{Training Reward Curve}
    \label{fig:train_reward}
  \end{subfigure}
  \hfill
  \begin{subfigure}[b]{0.32\textwidth}
    \includegraphics[width=\textwidth]{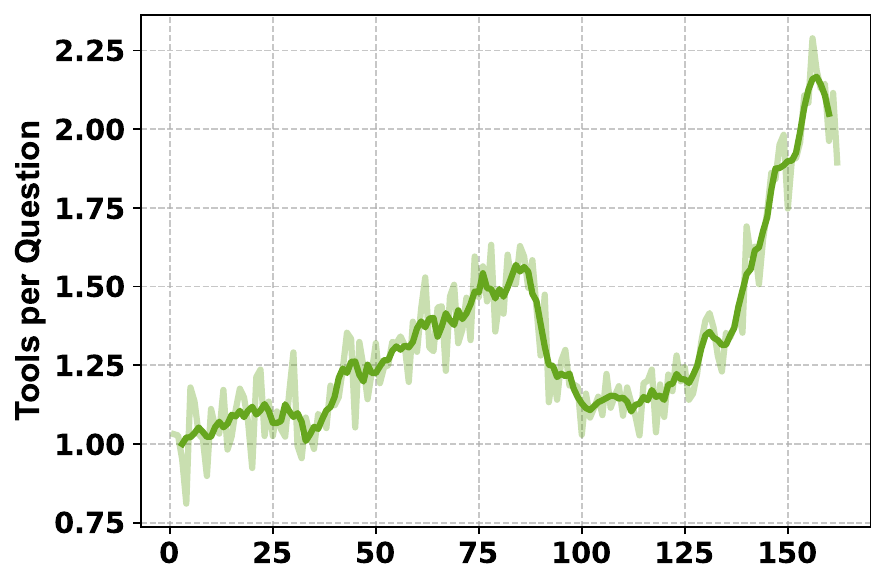}
    \caption{\#Tools per Question}
    \label{fig:tools_per_ques}
  \end{subfigure}
  \caption{
    Analysis of our RL training process. The $x$-axis represents training steps.
    (a) HLE score progression (evaluated on 320 random questions) shows consistent improvement from $\sim$2\% to over 10\%.
    (b) Training reward stabilizes around 0.4 after initial increase.
    (c) Tool usage per question increases from $\sim$1 to over 2, indicating that \modelname learns to leverage multiple tools for complex questions.
    Extended analysis including tool selection preferences and response length dynamics is provided in Appendix~\ref{app:rl_extended}.
  }
  \label{fig:rl_process}
\end{figure*}

%% file: tables/ablation_tool.tex
\begin{table*}[t]
\centering
\setlength{\tabcolsep}{4pt}
\renewcommand{\arraystretch}{1.2}
\caption{Ablation Study on Tools for HLE.}
\label{tab:ablation_tool} 
\small
\resizebox{\linewidth}{!}{
\begin{tabular}{@{}cccccccccccc@{}}
\toprule
\multicolumn{3}{c}{\textbf{Tools}}                 & \multicolumn{9}{c}{\textbf{Humanity’s Last Exam}}                                                                                                                                \\ \midrule
\python Python & \google Search & \scholar Scholar & \textbf{Bio/Med}  & \textbf{Chem.}   & \textbf{CS/AI}   & \textbf{Engineering} & \textbf{Humanities} & \textbf{Math}    & \textbf{Physics} & \textbf{Other}   & \textbf{Avg.}    \\ \midrule
\cmark         & \cmark         & \cmark           & \underline{12.66} & 3.00             & \underline{5.75} & \underline{4.83}     & \textbf{11.92}      & \textbf{6.46}    & \textbf{6.43}    & \underline{7.42} & \textbf{7.38}    \\ \midrule
\xmark         & \cmark         & \cmark           & \textbf{13.12}    & 5.00             & 5.31             & \underline{4.83}     & 8.29                & 4.92             & 5.45             & 5.14             & \underline{6.21} \\
\cmark         & \xmark         & \cmark           & 11.31             & \underline{6.00} & \textbf{7.07}    & 3.22                 & 6.73                & 4.92             & 3.47             & 6.85             & 5.99             \\
\cmark         & \cmark         & \xmark           & 11.76             & \textbf{7.00}    & 4.42             & \underline{4.83}     & \underline{8.81}    & \underline{6.05} & 4.45             & \textbf{8.00}    & 6.72             \\ \midrule
\cmark         & \xmark         & \xmark           & 4.54              & 4.00             & 3.98             & \textbf{6.45}        & 4.14                & 5.02             & 3.96             & 4.57             & 4.64             \\
\xmark         & \cmark         & \xmark           & \underline{12.66} & \underline{6.00} & 5.31             & 3.22                 & 7.77                & 4.41             & \underline{5.94} & \textbf{8.00}    & 6.12             \\
\xmark         & \xmark         & \cmark           & 12.21             & \textbf{7.00}    & 4.42             & \underline{4.83}     & 5.69                & 4.51             & 5.44             & 4.57             & 5.61             \\ \bottomrule
\end{tabular}
}
\end{table*}

%% file: tables/pages_per_ques.tex

\begin{table}
    \centering
    \caption{Information Volume per Question}
    \label{tab:pages_per_ques} 
    \small
    \begin{tabular}{@{}lccc@{}}
    \toprule
    \textbf{Dataset} & \textbf{\google Web Pages} & \textbf{\scholar Papers} & \textbf{\python Python} \\ \midrule
    PopQA            & 18.81                      & 0.04                     & 0.0                     \\
    hotpotqa         & 17.38                      & 0.12                     & 0.002                   \\
    HLE              & \textbf{22.31}             & \textbf{0.17}            & \textbf{0.05}           \\ \bottomrule
    \end{tabular}
\end{table}

%% file: section/5_conclusion.tex
\section{Conclusion}
We presented \modelname, a co-evolution framework that jointly trains System 1 (efficient information processing) and System 2 (deliberate reasoning) via multi-agent RL, with decoupled gradients, bin-packing, and advantage-balanced sampling to enable effective collaboration. Under a Zero RL setting (no SFT), \modelname (8B) achieves 8.17\% on HLE and improves performance by 8.9\% on average across 7 knowledge-intensive QA tasks; ablations (e.g., removing System 1) confirm that the gains come from learned co-evolution rather than tool access alone.

\clearpage

%% file: section/6_appendix.tex
\newpage
\appendix




\section{Additional Related Work}\label{app:related_work}

\paragraph{Language Reasoning Models (LRMs).}
Recent advances in language models have witnessed the emergence of LRMs, which specifically excel at deliberate, System 2-type thinking~\citep{jaech2024openai,yu2024distilling,guo2025deepseek,qwen_qwq,claude_37,li2025system,ziabari2025reasoning,chen2025reform}.
Despite these advances, current LRMs rely on static, parameterized knowledge acquired during pre-training, without access to external world information~\citep{c3po2025,li2025search,jin2025search,li2025webthinker}.
Furthermore, when faced with simpler tasks, such as analyzing web pages or reading papers, LRMs may tend to overanalyze, consuming excessive reasoning tokens that limit their capacity to process large volumes of information. Our work addresses these limitations by integrating System 1's fast, intuitive thinking with System 2's deliberate reasoning in a multi-agent framework that leverages external knowledge sources.

\paragraph{Reasoning Models with External Tools.}
Recent work has explored integrating external tools into reasoning models~\citep{chen2024alphamath,chen2024step,li2025search,jin2025search,li2025webthinker,c3po2025,chen2025iterresearch}.
Search-R1~\citep{jin2025search} and Search-o1~\citep{li2025search} directly feed raw search results to reasoning models, which can overwhelm the context with irrelevant information.
WebThinker~\citep{li2025webthinker} relies on a \textit{fixed} extraction prompt to process retrieved documents before reasoning, while C-3PO~\citep{c3po2025} uses supervised fine-tuning with curated trajectories but does not co-evolve the information processor and reasoner.
In contrast, \modelname enables \textbf{co-evolution} between System 1 and System 2 through multi-agent RL, allowing System 1 to learn \textit{task-specific} information distillation that adapts to System 2's needs.

\paragraph{Multi-agent Systems.}
Multi-agent systems have gained significant attention in the LLM community as a means to tackle complex tasks through collaborative problem-solving~\citep{park2023generative, wu2023autogen,li2023camel,hong2023metagpt, qian2023communicative,han2024llm,li2024survey,c3po2025}. Recent work has explored various agent architectures, from simple role-playing approaches~\citep{li2023camel, du2023improving} to more sophisticated frameworks with specialized agents handling different aspects of complex tasks~\citep{hong2023metagpt, wang2023voyager}.
However, most existing multi-agent systems employ agents with similar cognitive architectures, typically all operating in a intuitive, System1-type thinking or deliberate, System 2-like reasoning mode. This homogeneity limits their ability to efficiently process large volumes of information while maintaining complex reasoning capabilities. Our approach diverges by implementing a heterogeneous multi-agent system where System 1 and System 2 agents possess complementary cognitive abilities, optimized through reinforcement learning~\citep{abs-2312-01058,ZhuDW24,shao2024deepseekmath} to collaboratively tackle complex research tasks requiring both efficient information processing and deliberate reasoning.




\section{Training Data Filtering Pipeline}\label{app:data_filter}

\begin{figure}[h]
    \begin{center}
    \includegraphics[width=\linewidth]{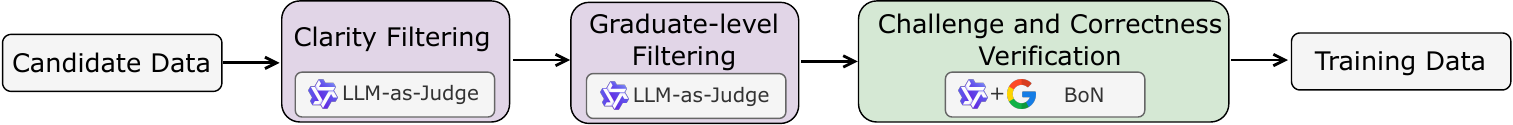}
    \caption{Our data curation pipeline.}
    \label{fig:data_pipeline}
    \end{center}
\end{figure}

Given the absence of specialized training datasets for enhancing model capabilities required for challenging benchmarks such as HLE~\citep{phan2025humanity} (which demands web browsing, reasoning, and computation skills), we propose a data curation pipeline that selectively identifies suitable training examples from open-source datasets. 
As illustrated in Figure~\ref{fig:data_pipeline}, our pipeline evaluates candidate examples based on three key dimensions: clarity of presentation, problem complexity, and solution correctness.

\begin{figure}[t]
    \begin{center}
    \includegraphics[width=0.7\linewidth]{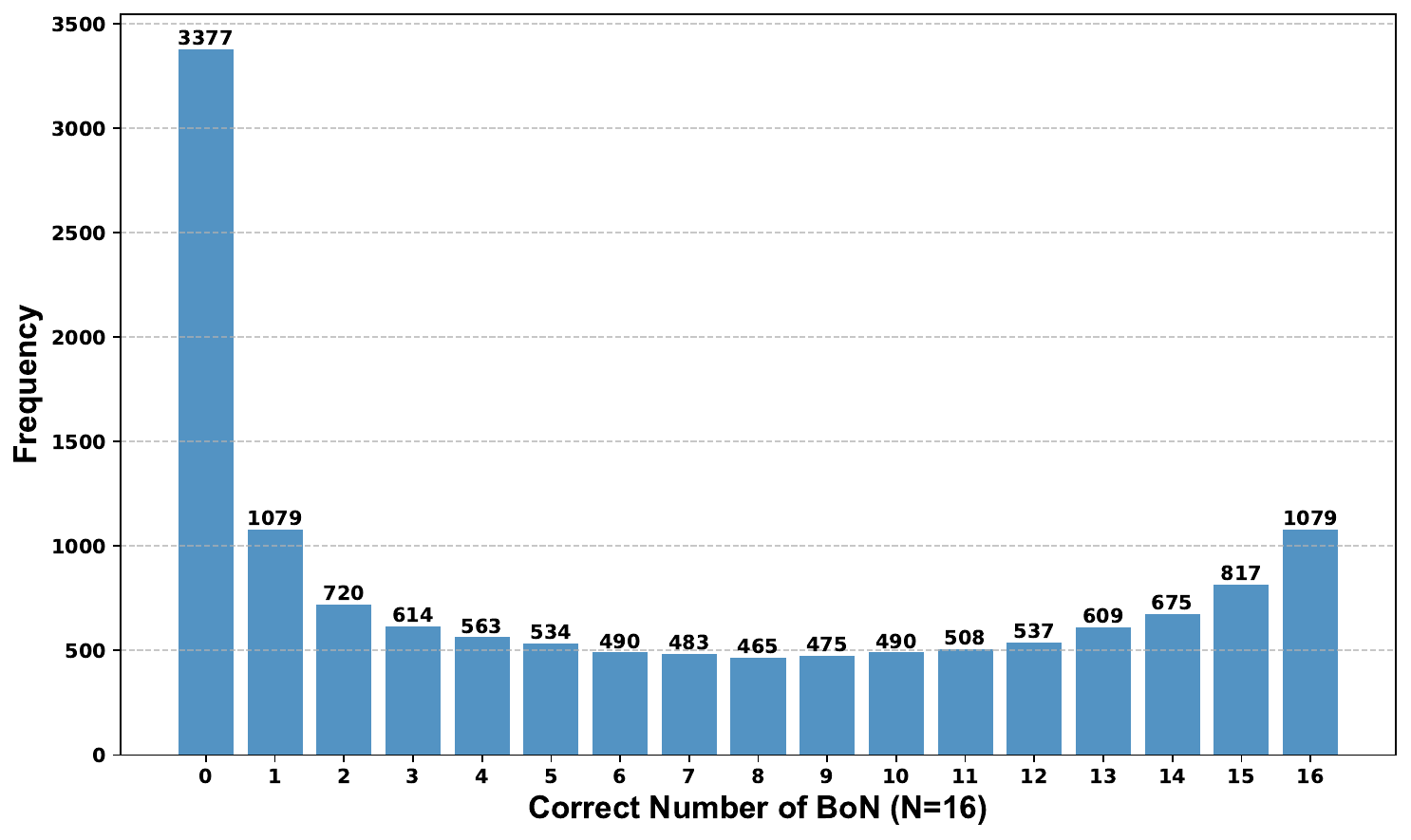}
    \caption{Distribution of Correct Number in Best-of-N ($N=16$). Questions answered correctly 1-12 times were retained for training, while those answered 0 times (potentially ambiguous or lacking definitive solutions) or >12 times (too trivial) were excluded.}
    \label{fig:data_bon}
    \end{center}
\end{figure}

All candidate data are sourced from publicly available datasets, including MMIQC~\cite{liu2025augmenting}, WebInstructSub~\cite{yue2024mammoth2}, GeneralThought\footnote{\url{https://huggingface.co/datasets/GeneralReasoning/GeneralThought-430K}}, and SuperGPQA~\cite{du2025supergpqa}. 
To ensure quality for our dual-system training, we implement a rigorous multi-stage filtering process:
\begin{enumerate}[topsep=1pt, partopsep=1pt, leftmargin=12pt, itemsep=-1pt]
    \item \textbf{Initial filtering by academic level:} Starting from an initial pool of 5 million examples, we categorize each prompt by academic discipline and retain only those that may meet undergraduate or graduate-level difficulty standards, reducing the dataset to 237K examples.
    \item \textbf{Deduplication:} We remove near-identical prompts that differ only in phrasing or superficial elements, yielding 155K unique prompts.
    \item \textbf{Clarity assessment:} We employ an LLM (Qwen2.5-72B-Instruct~\citep{yang2024qwen2}) to systematically evaluate the clarity of each prompt, filtering out those deemed ambiguous or poorly formulated, resulting in 99K high-quality prompts.
    \item \textbf{Graduate-level Filtering:} We further use the same LLM to assess the challenge level of each prompt, retaining only those that meet graduate-level standards in their respective disciplines, resulting in 81K prompts of appropriate difficulty.
    \item \textbf{Challenge and Correctness Verification:} To ensure both appropriate difficulty and answer verifiability, we perform a best-of-16 (BoN) sampling procedure using Qwen2.5-72B-Instruct with Google Search access. As shown in Figure~\ref{fig:data_bon}, we plot the distribution of how many times (out of 16 attempts) the model correctly answers each question. We retain only questions with moderate difficulty (correctly answered 1-12 times out of 16 attempts), which serves two critical purposes: (1) eliminating trivial questions (answered correctly >12 times) and excessively difficult ones (never answered correctly), and (2) identifying questions with consistent, verifiable answers, as those never answered correctly may lack definitive solutions or contain ambiguous information.
    \item The final dataset comprises 40K carefully curated prompts spanning diverse academic disciplines and difficulty levels, specifically designed to support high-quality dual-system training for complex reasoning tasks.
\end{enumerate}
 
To facilitate research advancement and reproducibility, we have open-sourced our curated 40K dataset. The detailed prompts used throughout our data curation pipeline are provided in Appendix~\ref{app:prompt_data}.



\section{More Implementation Details}\label{app:imple_details}

In this section, we provide a comprehensive implementation details of our proposed method. For additional insights and more intricate details, we refer the reader to our supplementary materials.

\subsection{Overall Algorithm}
\input{algorithm/overall}
Our dual-system approach leverages the complementary strengths of two distinct cognitive systems within the same large language model. Algorithm \ref{alg:rollot} presents the complete multi-agent rollout process that orchestrates the interaction between System 1 (fast, intuitive processing) and System 2 (deliberate, analytical reasoning).

Inspired by human cognition, our \modelname assigns distinct roles to each system: System 2 manages strategic reasoning and decision-making, while System 1 efficiently processes and distills large volumes of information. The algorithm enables System 2 to independently continue reasoning when no tool is required, and terminates either when an answer is reached or after the maximum number of turns is exhausted.
The parallel processing capability of System 1 is particularly valuable when handling extensive tool-retrieved information. By distributing information processing across multiple parallel instances of System 1, our approach efficiently manages complex information needs without exceeding context window limitations.

\subsection{Bin-Packing Details}\label{app:bin_packing}

As mentioned in Section~\ref{sec:bin_packing}, we employ a First Fit Decreasing (FFD) algorithm to efficiently organize variable-length tool outputs into optimally-sized chunks. Our bin-packing implementation follows these key steps:
\begin{enumerate}[topsep=1pt, partopsep=1pt, leftmargin=12pt, itemsep=-1pt]
    \item \textbf{Token counting}: For each tool output $o_{t_i}^{(j)}$, we count the number of tokens using the model's tokenizer.
    \item \textbf{Large output handling}: If any single output exceeds the maximum context length of System 1, it is truncated and placed in a dedicated bin.
    \item \textbf{Sorting}: Remaining outputs are sorted in decreasing order of their token lengths.
    \item \textbf{Bin assignment}: Each output is assigned to the first bin that can accommodate it without exceeding the context length limit. If no existing bin has sufficient space, a new bin is created.
\end{enumerate}

\subsection{Reward Design}

We introduce a straightforward yet effective reward design for our multi-agent RL training. For each rollout trajectory, we employ LLMs (Qwen2.5-72B-Instruct) as evaluators to assess the correctness of the predicted final answer following HLE's official prompt~\citep{phan2025humanity}.
The reward function is defined as:
\begin{equation}
r(c_N, \text{ground truth}) = 
\begin{cases}
1, & \text{if } \text{Eval}_{\text{LLM}} = \text{Correct} \\
0, & \text{otherwise}
\end{cases}
\end{equation}
where $c_N$ is the final reasoning trajectory, $N$ is the number of interaction turns. We will extract the answer part between \texttt{<answer>} and \texttt{</answer>} tags for evaluation.

All System 1 and System 2 samples within the same trajectory share this trajectory-level reward. This design reflects the inherently collaborative nature of our dual-system framework, where the final answer quality depends on both System 2's reasoning and System 1's information processing. By sharing rewards, we encourage both systems to optimize toward the same goal—producing correct final answers—rather than pursuing potentially conflicting individual objectives.
This binary reward signal creates a clear learning objective for both systems: System 2 learns to generate better reasoning steps and more effective tool-use plans, while System 1 learns to distill information more accurately and concisely to support System 2's reasoning. The simplicity of this reward function helps avoid the common pitfalls of overly complex reward engineering while maintaining focus on the ultimate goal of correct problem-solving.



\subsection{RL Loss}\label{app:grpo}

We write the policy loss of one single rollout trajectory as follows.
\begin{equation}
    \mathcal{L}_{\text{policy}}(x, y, A_{\text{sys}_i}) = \frac{1}{|y|}\sum_{j=1}^{|y|}\min\left[\frac{\pi_{\text{sys}_i}(y_j|x, y_{<j})}{\pi_{\text{sys}_i}^{\text{old}}(y_j|x, y_{<j})} A_{\text{sys}_i}, \text{clip}\left( \frac{\pi_{\text{sys}_i}(y_j|x, y_{<j})}{\pi_{\text{sys}_i}^{\text{old}}(y_j|x, y_{<j})}, 1-\epsilon, 1+\epsilon \right) A_{\text{sys}_i}\right]
\end{equation}
where $y_j$ is the $j$-th token of LLM output $y$. 
Similarly, we write the KL loss as follows.
\begin{equation}
    \mathcal{L}_{\text{KL}}(x, y) = - \frac{1}{|y|} \sum_{j=1}^{|y|} \mathbb{D}_{\text{KL}}\left(\pi_{\text{sys}_i}(y_j|x, y_{<j}) \| \pi_{\text{sys}_i}^{\text{ref}}(y_j|x, y_{<j}) \right)
\end{equation}
In practical implementation, the expectation in Eq.~(\ref{eq:grpo}) is achieved via averaging over a group of $G$ rollouts as well as a batch training examples.

\subsection{Implementation Details}
\input{tables/rl_params}
This section provides comprehensive implementation details of our \modelname.
We initialize our policy model with Qwen2.5-7B-Instruct~\citep{yang2024qwen2}, which serves as the foundation for both System 1 and System 2.
Furthermore, we provide external tools such as Google Search, Google Scholar via the SerpAPI\footnote{\url{https://serpapi.com/}}, and Python code interpreter for supporting autonomous tool selection during the reasoning process.
We employ the GRPO algorithm~\citep{shao2024deepseekmath} with the following hyperparameters: learning rate of 1e-6, batch size of 32, sampled responses per prompt (group size $G$) of 16, temperature of 1.0, top-p of 0.95, and both KL loss coefficient and entropy coefficient set to 0.
Additionally, we set different maximum lengths for System 1 and System 2: prompt lengths of 23,552 and 3,072 tokens, and response lengths of 8,192 and 28,672 tokens, respectively. This configuration enables System 1 to better incorporate external knowledge while allowing System 2 to focus on sophisticated multi-step reasoning.
For all baselines, we maintain their original settings in their respective papers to ensure optimal performance.
Table~\ref{tab:rl_params} summarizes the key hyperparameters used during the reinforcement learning phase.

The asymmetric prompt and response length configurations between System 1 and System 2 are designed to leverage their complementary roles. System 2, with its shorter prompt length (3,072 tokens) but longer response capability (28,672 tokens), is optimized for detailed reasoning and solution generation. Conversely, System 1, with its extended prompt length (23,552 tokens) but more concise response limit (8,192 tokens), excels at processing and summarizing large volumes of information. This configuration aligns with cognitive science theories where System 2 handles deliberate reasoning while System 1 processes information rapidly.

\subsection{Baseline Comparisons and Fairness}\label{app:baseline_fair}

We provide detailed clarifications on the fairness of our experimental comparisons, addressing potential concerns about model size, training methodology, and tool availability.

\paragraph{Zero RL Setting.}
A critical distinction of \modelname is that we adopt a ``Zero RL'' training paradigm: we initialize from the base model (Qwen2.5-7B-Instruct or Qwen3-8B) and train \textit{directly with RL without any Supervised Fine-Tuning (SFT)}. This is a significantly more challenging setting than approaches that use SFT warmup:
\begin{itemize}[topsep=1pt, partopsep=1pt, leftmargin=8pt, itemsep=0pt]
    \item \textbf{WebThinker} uses SFT on curated trajectories before RL training, providing a strong initialization.
    \item \textbf{C-3PO} relies on SFT with a 72B model, leveraging both scale and supervised data.
    \item \textbf{\modelname} trains from scratch with RL only, yet outperforms these larger, SFT-trained models.
\end{itemize}
This demonstrates that our multi-agent RL framework, rather than model scale or supervised data quality, is the key driver of performance.

\paragraph{Model Size Comparison.}
We use 7B/8B models, significantly smaller than WebThinker (32B) and C-3PO (72B). The fact that \modelname outperforms these larger models despite the parameter disadvantage further validates our architectural and algorithmic contributions.

\paragraph{Data Source Parity.}
Our training data is curated exclusively from public sources (Appendix~\ref{app:data_filter}), including MMIQC, WebInstructSub, GeneralThought, and standard QA datasets. We do not use proprietary data or expensive human annotations. This ensures fair comparison with baselines that also use public data sources.

\paragraph{Tool Availability.}
\modelname uses three tools: Google Search, Google Scholar, and Python Interpreter. Some baselines (e.g., WebThinker) do not use Google Scholar. This difference stems from architectural constraints: baselines without a learned information processor cannot effectively handle the large volume of academic papers that Scholar returns. Our System 1 can efficiently distill this information, enabling effective use of Scholar. Importantly, as shown in our ablation (Section 3.6), the performance gains come from our co-evolution framework, not merely tool availability—the ``w/o System 1'' setting maintains all tools but shows significant performance degradation.

\paragraph{WebThinker Score Discrepancy.}
Readers may notice that our reported WebThinker results differ from the original paper~\citep{li2025webthinker}. This stems from two factors:
\begin{enumerate}[topsep=1pt, partopsep=1pt, leftmargin=12pt, itemsep=-1pt]
    \item \textbf{Evaluation scale}: We evaluate on the complete HLE text-based question set (2,154 questions), rather than the 500-question sample used in the original paper.
    \item \textbf{Evaluation criteria}: We strictly use HLE's official evaluation prompt for all models. The original WebThinker paper used a custom prompt that resulted in more lenient scoring.
\end{enumerate}

\paragraph{Why Share a Single Checkpoint for System 1 and System 2?}
Our design choice to implement both systems within a single LLM (activated through different prompts) is deliberate and motivated by several factors:
\begin{itemize}[topsep=1pt, partopsep=1pt, leftmargin=8pt, itemsep=0pt]
    \item \textbf{Alignment with dual-process theory}: In cognitive science, System 1 and System 2 are not separate brains but different \textit{modes} of the same mind. Our single-model design reflects this—one model that can operate in both modes.
    \item \textbf{Synergistic optimization}: Sharing parameters enables tighter co-adaptation. System 2 learns to generate purposes that System 1 (sharing the same knowledge) can effectively interpret, and vice versa.
    \item \textbf{Practical efficiency}: A two-model design would double memory requirements and inference complexity, limiting practical deployment.
    \item \textbf{Empirical validation}: WebThinker's two-component design (separate reasoning model + fixed extractor) underperforms our unified approach despite using a larger model (32B vs. 8B), suggesting our single-model co-evolution is more effective.
\end{itemize}

\subsection{Dataset Details}
Our training data consists of two complementary components: (1) a curated collection of complex reasoning examples filtered through our specialized pipeline, as shown in Appendix~\ref{app:data_filter}, and (2) several established open-source single-hop, multi-hop, and Biology \& Medicine datasets that enhance the model's knowledge retrieval and reasoning capabilities.
Table~\ref{tab:data_statistics} summarizes the key statistics of our complete training dataset.
\input{tables/data}

As shown in Table~\ref{tab:data_statistics}, we randomly sampled a total of 5,050 training examples across eight distinct datasets. The composition is carefully balanced to ensure comprehensive coverage of different reasoning types and knowledge domains.

\begin{itemize}[leftmargin=*, noitemsep, topsep=0pt]
    \item \textbf{Our curated data (2,000 examples, 39.6\%):} These examples were selected from our pipeline-filtered corpus described in Appendix~\ref{app:data_filter}, which contains high-quality, graduate-level complex reasoning tasks vetted through our rigorous multi-stage filtering process.
    
    \item \textbf{Single-Hop QA (800 examples, 15.8\%):} We incorporated 400 randomly sampled examples each from TriviaQA~\citep{JoshiCWZ17} and PopQA~\citep{MallenAZDKH23}. These datasets focus on direct factual knowledge retrieval, with TriviaQA covering a broad range of trivia questions and PopQA specifically targeting popular entities and common knowledge.
    
    \item \textbf{Multi-Hop QA (1,500 examples, 29.7\%):} To strengthen the model's multi-step reasoning capabilities, we included 500 examples each from HotpotQA~\citep{yang-etal-2018-hotpotqa}, 2WikiMultihopQA \citep{HoNSA20}, and MuSiQue \citep{trivedi-etal-2022-musique}. These datasets require reasoning across multiple documents or knowledge pieces to arrive at the correct answer.
    
    \item \textbf{Biology \& Medicine (750 examples, 14.9\%):} To enhance domain-specific knowledge, we sampled 500 examples from PubMedQA~\citep{reese2024evaluation} and 250 examples from CUPCase~\citep{perets2025cupcase}, covering biomedical research questions and clinical case analysis respectively.
\end{itemize}

\subsection{Experiment Environments}
All experiments were conducted on Ubuntu 22.04 equipped with NVIDIA A100 GPUs. Our implementation relies on Python 3.10\footnote{\url{https://www.python.org/}} and PyTorch 2.6.0\footnote{\url{https://pytorch.org/}}, while extending VeRL\footnote{\url{https://github.com/volcengine/verl}} for our multi-agent reinforcement learning framework. For efficient execution, we implemented rollout procedures based on Qwen-Agent\footnote{\url{https://github.com/QwenLM/Qwen-Agent}} and use vLLM\footnote{\url{https://github.com/vllm-project/vllm}} and SGLang\footnote{\url{https://github.com/sgl-project/sglang}} as our inference engines.
We use Code Sandbox\footnote{\url{https://github.com/bytedance/SandboxFusion}} for Python Interpreter.

\section{Additional Experimental Results}\label{app:additional_exp}

\subsection{Extended RL Training Analysis}\label{app:rl_extended}

\begin{figure}[h]
  \centering
  \begin{subfigure}[b]{0.32\textwidth}
    \includegraphics[width=\textwidth]{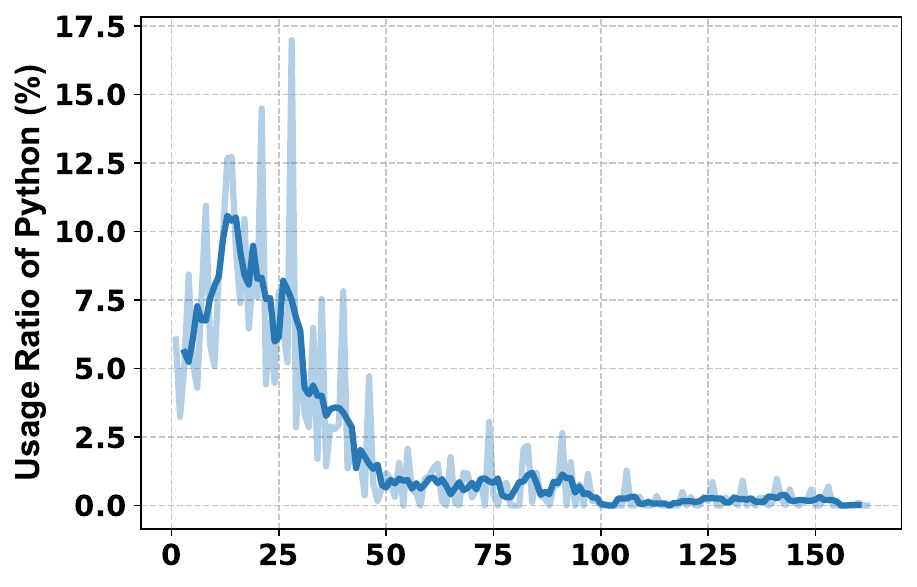}
    \caption{Usage Ratio of Python}
    \label{fig:python_ratio_app}
  \end{subfigure}
  \hfill
  \begin{subfigure}[b]{0.32\textwidth}
    \includegraphics[width=\textwidth]{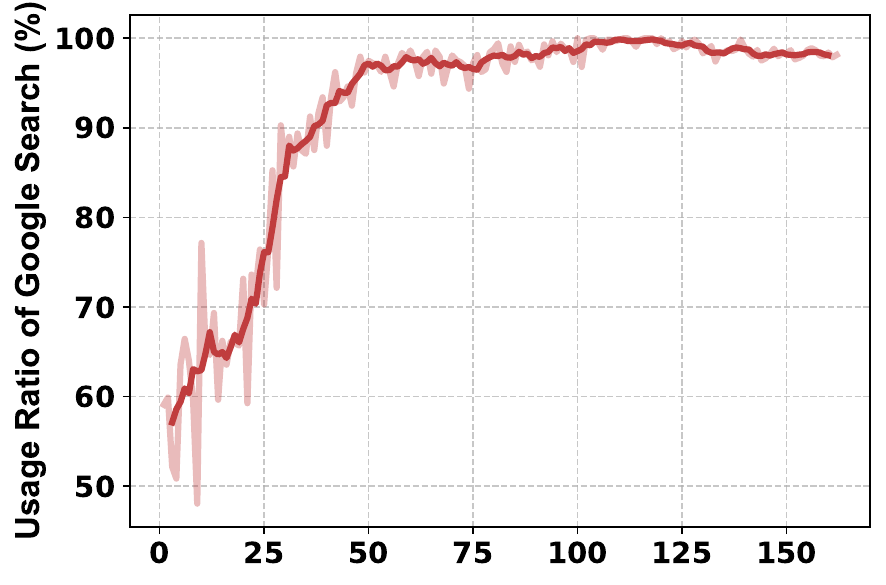}
    \caption{Usage Ratio of Google Search}
    \label{fig:google_ratio_app}
  \end{subfigure}
  \hfill
  \begin{subfigure}[b]{0.32\textwidth}
    \includegraphics[width=\textwidth]{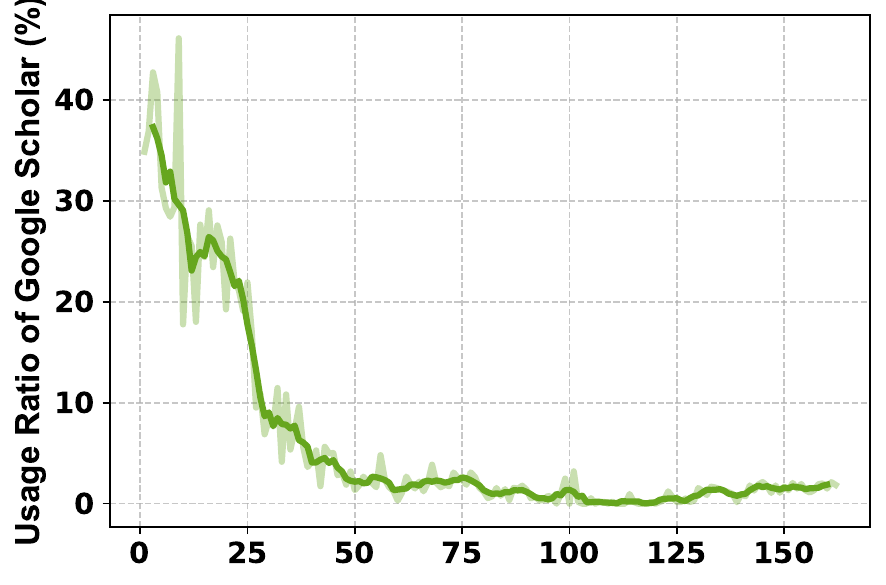}
    \caption{Usage Ratio of Google Scholar}
    \label{fig:schoar_ratio_app}
  \end{subfigure}

  \begin{subfigure}[b]{0.32\textwidth}
    \includegraphics[width=\textwidth]{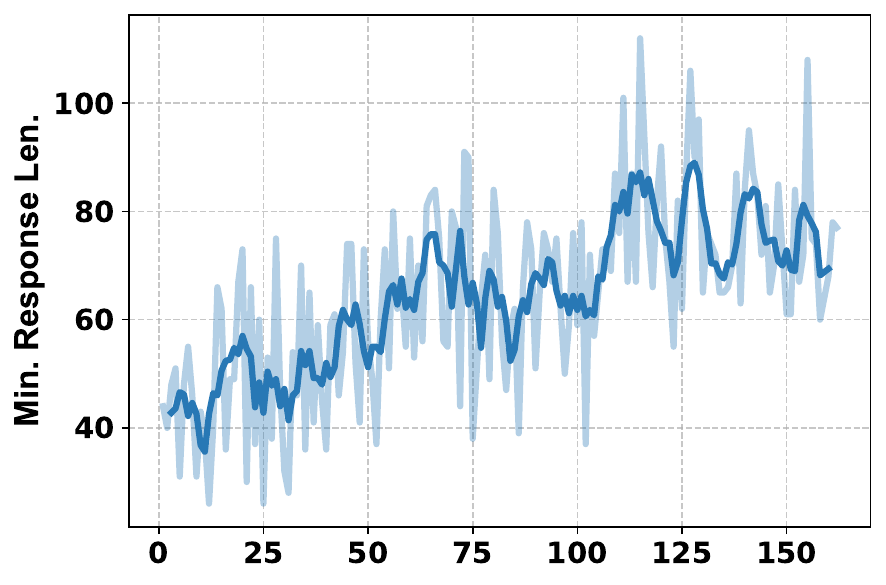}
    \caption{Min. Response Len. (Sys. 1)}
    \label{fig:resposne_min_app}
  \end{subfigure}
  \hfill
  \begin{subfigure}[b]{0.32\textwidth}
    \includegraphics[width=\textwidth]{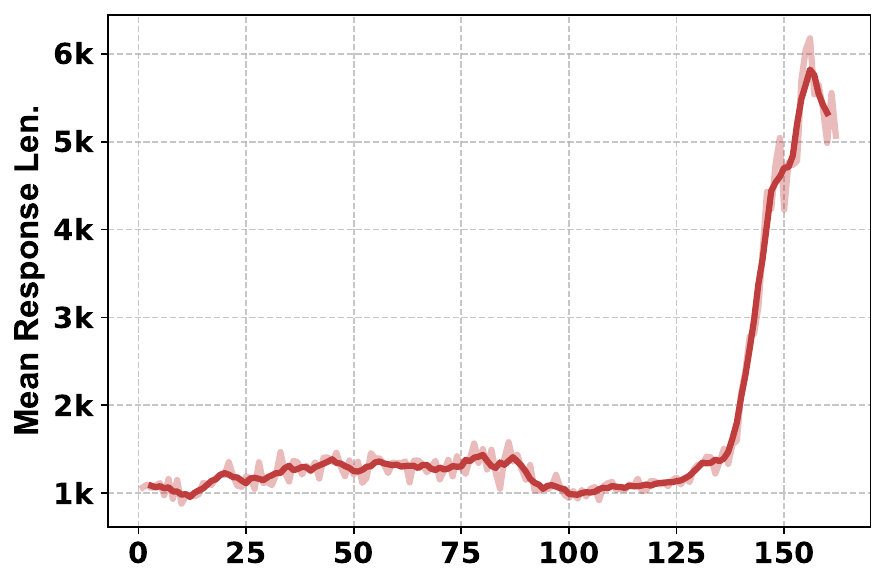}
    \caption{Mean Response Len.}
    \label{fig:response_mean_app}
  \end{subfigure}
  \hfill
  \begin{subfigure}[b]{0.32\textwidth}
    \includegraphics[width=\textwidth]{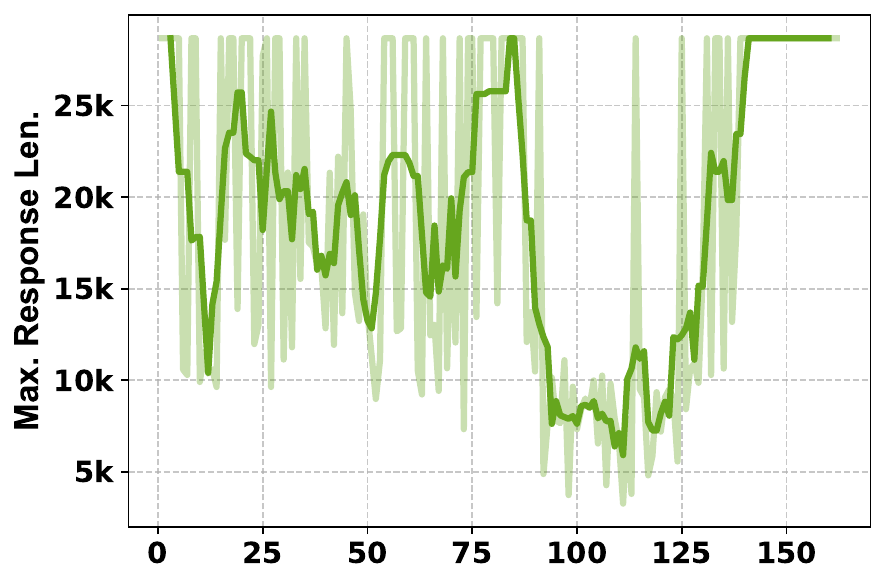}
    \caption{Max. Response Len. (Sys. 2)}
    \label{fig:response_max_app}
  \end{subfigure}
  \caption{Extended analysis of RL training dynamics. (a-c) Evolution of tool selection preferences. While Google Search emerges as the predominantly chosen tool, the model maintains capability to utilize Python and Google Scholar when appropriate. (d-f) Response length distributions showing that both systems learn to generate more comprehensive outputs as training advances.}
  \label{fig:rl_extended}
\end{figure}

Figure~\ref{fig:rl_extended}(a--c) shows how the learned policy allocates tool usage over training. Early in training, the model often relies on a narrow subset of tools (typically Google Search) and under-utilizes specialized tools, reflecting limited exploration and weak credit assignment under trajectory-level rewards. As training progresses, the tool distribution becomes more structured: Google Search remains the default high-recall tool, while Python and Scholar are invoked selectively when the question demands exact computation or academic evidence.

Figure~\ref{fig:rl_extended}(d--f) reports the evolution of response lengths. We observe that System 1's minimum response length increases over training, suggesting that it learns to consistently produce non-trivial, purpose-conditioned distillations (rather than empty or overly short summaries). Meanwhile, the mean response length increases and System 2's maximum response length grows, indicating that System 2 learns to sustain longer multi-step reasoning and to integrate evidence across multiple tool calls.

Together, these trends support two conclusions. First, the performance gains are not solely due to calling tools more frequently; instead, training improves the \emph{quality} of tool use (better timing, better specialization, and better evidence integration). Second, the co-evolution design stabilizes learning under noisy retrieval: System 2 can expand its search when needed, while System 1 limits context dilution by filtering irrelevant content and preserving task-critical entities/constraints.

\section{Instruction Templates}\label{app:prompt}

\subsection{Instruction of System 1 \& 2 in our \modelname}
\input{prompts/system1}

\input{prompts/system2}

\subsection{Instruction for our data curtion pipeline}\label{app:prompt_data}

In our data curation pipeline, we employed specific instructions for each filtering stage to ensure consistent evaluation criteria. Below, we provide the detailed prompts used for Clarity Filtering and Graduate-level Filtering stages, which were directly provided to Qwen2.5-72B-Instruct for assessment.

For the Challenge and Correctness Verification stage (Best-of-16 sampling), we utilized a different approach. Rather than using a standalone evaluation prompt, we leveraged our dual-system framework itself, combining the System 1 and System 2 instructions (detailed in Appendix~\ref{app:prompt}) with Google Search integration. This allowed Qwen2.5-72B-Instruct to perform complete reasoning attempts on each question, providing a more authentic assessment of difficulty level and answer verifiability than a static evaluation would allow.

\input{prompts/data}

\subsection{Instruction for Reward Model}
For our reward model, we strictly adhere to the official evaluation criteria from the Humanity's Last Exam (HLE) benchmark \citep{phan2025humanity}.

\input{prompts/reward}

%% file: algorithm/overall.tex
\begin{algorithm}[t]
\caption{Multi-agent Rollout Process}\label{alg:rollot}
\begin{algorithmic}[1]
\REQUIRE Question $q$, System 1 $\pi_{\text{sys}_1}$ and System 2 $\pi_{\text{sys}_2}$ in the same LLM, candidate tools, Maximum interaction turns $N_{\max}$.
\STATE $i\leftarrow0$ \COMMENT{Initialize interaction turns}
\STATE $c_0 \leftarrow \{q\}$ \COMMENT{Initialize reasoning context for System 2}
\STATE $c_{\text{sys}_1} \leftarrow \emptyset$ \COMMENT{Initialize collection of System 1 input-output pairs}
\WHILE{$i < N_{\max}$}
    \STATE $s_i, (t_i, p_i) \leftarrow \pi_{\text{sys}_2}(c_i)$ \COMMENT{System 2 generates reasoning step $s_i$, optional tool request $t_i$ and purpose $p_i$}
    \IF{$t_i$ and $p_i$ are not Empty}
        \STATE $\{o_{t_i}^{(1)}, o_{t_i}^{(2)}, \ldots, o_{t_i}^{(n_{t_i})}\} \leftarrow \text{Tool Call}(t_i)$ \COMMENT{Execute tool call}
        \STATE $\{b_1, \ldots, b_m\} \leftarrow \text{Bin-Packing}(o_{t_i}^{(1)}, \ldots, o_{t_i}^{(n_{t_i})})$ \COMMENT{$m$ chunks}
        \STATE $\{d_1, \ldots, d_m\} \leftarrow \{\pi_{\text{sys}_1}(b_1), \ldots, \pi_{\text{sys}_1}(b_m)\}$ \COMMENT{Parallel distillation}
        \STATE $c_{\text{sys}_1} \leftarrow c_{\text{sys}_1} \cup \{(b_j, d_j)\}_{j=1}^m$ \COMMENT{Collect input-output pairs}
        \STATE $\tilde{o}_{t_i} \leftarrow \cup_{j=1}^m d_j$ \COMMENT{Combine distilled information}
        \STATE $c_{i+1} \leftarrow c_i \cup \{s_i, t_i, p_i, \tilde{o}_{t_i}\}$ \COMMENT{Update context with tool results}
    \ELSE
        \STATE $c_{i+1} \leftarrow c_i \cup \{s_i\}$ \COMMENT{Update context with reasoning only}
        \IF{answer tag detected in $s_i$}
            \STATE \textbf{break} \COMMENT{Answer provided}
        \ENDIF
    \ENDIF
    \STATE $i \leftarrow i + 1$
\ENDWHILE
\STATE \textbf{return} Rollout trajectory $c_{i}$ and System 1 input-output pairs $c_{\text{sys}_1}$
\end{algorithmic}
\end{algorithm}

%% file: tables/rl_params.tex
\begin{table}[h]
\centering
\caption{Key hyperparameters in the RL phase.}
\label{tab:rl_params} 
\begin{tabular}{@{}lc@{}}
\toprule
\textbf{Hyperparameter}       & \textbf{Value} \\ \midrule
Learning Rate of Policy model & 1e-6           \\
Base model  & Qwen2.5-7B-Instruct           \\
Batch size                    & 32           \\
$G$               & 16         \\
temperature                & 1.0           \\
KL loss coefficient $\lambda$                & 0.           \\
entropy coefficient                & 0.           \\
Maximum Prompt Length of System 1              & 23,552           \\
Maximum Response Length of System 1              & 8192           \\
Maximum Prompt Length of System 2              & 3072           \\
Maximum Response Length of System 2              & 28,672           \\
Maximum interaction turns                 & 10             \\ \bottomrule
\end{tabular}
\end{table}

%% file: tables/data.tex
\begin{table}[h]
\vspace{-1em}
\centering
\setlength{\tabcolsep}{4pt}
\renewcommand{\arraystretch}{1.2}
\caption{Data Statistics.}
\label{tab:data_statistics} 
\resizebox{\linewidth}{!}{
\begin{tabular}{@{}l|c|cc|ccc|cc|c@{}}
\toprule
\textbf{Data Type}      & \multirow{2}{*}{\textbf{Our curated data}} & \multicolumn{2}{c|}{\textbf{Single-Hop QA}} & \multicolumn{3}{c|}{\textbf{Multi-Hop QA}} & \multicolumn{2}{c|}{\textbf{Biology\&Medicine}} & \multirow{2}{*}{\textbf{Total}} \\
\textbf{Data Name}      &                                            & TriviaQA               & PopQA              & HotpotQA       & 2Wiki      & Musique      & PubMedQA                & CUPCase               &                                 \\ \midrule
\textbf{Sampled Number} & 2000                                       & 400                    & 400                & 500            & 500        & 500          & 500                     & 250                   & 5050                            \\ \bottomrule
\end{tabular}
}
\vspace{-1em}
\end{table}

%% file: prompts/system1.tex
\begin{tcolorbox}[title=Instruction for System 1,width=\linewidth, breakable]
\textcolor{red}{\texttt{\textless{}System Prompt\textgreater{}}}\\
You are an expert information extractor. Your sole task is to extract only the information that directly supports the tool call's purpose or answers the user's question.\\

\#\# Task Guidelines\\
1. **Match Each Query**: For every query, extract information directly relevant to it and record its source (e.g., title, section name).\\
2. **Content Scanning**: Locate the **specific sections/data** directly related to the user's goal within the content.\\
3. **Key Extraction**: Identify and extract the **most relevant information** from the content, you never miss any important information\\
4. **Verbatim Key Content**: Preserve the original wording of key definitions, claims, formulas, data points.\\
5. **Preserve Detail**: Include relevant data, numbers, metrics, or formulas.\\
6. **Output Structure**: Organize the extracted content per query in a clear and nested way.\\

\textcolor{red}{\texttt{\textless{}User Prompt\textgreater{}}}\\
\textcolor{blue}{\{tool outputs\}}\\

\# User Question: \textcolor{blue}{\{question\}}

\end{tcolorbox}

%% file: prompts/system2.tex
\begin{tcolorbox}[title=Instruction for System 2,width=\linewidth, breakable]
\textcolor{red}{\texttt{\textless{}System Prompt\textgreater{}}}\\
You are an expert researcher who combines rigorous analytical reasoning with thorough information seeking abilities. You excel at solving complex problems through logical thinking, careful analysis, and responsible tool use. You are known for your careful and thorough approach, never rushing to conclusions without complete analysis.\\

\{\textcolor{blue}{tool description}\}\\

When performing a search:\\
1. **Persistent Actions for Answers**: You can engage in multiple search iterations, delving deeply into the topic to explore all possible aspects until a satisfactory answer is found.\\
2. **Repeated Verification**: Before presenting a Final Answer, you will **cross-check** and **validate the information** you've gathered to confirm its accuracy and reliability.\\
3. **Attention to Detail**: You will carefully analyze each information source to ensure that all data is current, relevant, and from credible origins.\\

Your reasoning process should be enclosed within <think> </think> tags. If you need external support, make tool calls inside <tool\_call> </tool\_call> tags. After a tool call, always reassess the result critically and continue your analysis in a new <think> section. Tools are helpful but not always reliable — treat their output with scrutiny.\\

Finally, present your key reasoning and final answer inside <answer> </answer> tags.

Do not nest tags. Each tag block must be independent.\\

\textcolor{red}{\texttt{\textless{}User Prompt\textgreater{}}}\\
\textcolor{blue}{\{question\}}

\end{tcolorbox}

%% file: prompts/data.tex
\begin{tcolorbox}[title=Instruction of Clarity Filtering,width=\linewidth, breakable]

\#\#\# Instruction \#\#\#\\
You are a domain expert proficient in various subjects such as Math, Physics, Biology, Humanities, Computer Science, Engineering, and Chemistry. You will be given a question and its corresponding answer. Please identify whether the given QA-pair exhibits "Clarity". Only output 1 for yes and 0 for no. \\

\#\#\# Definition \#\#\#\\
"Clarity" is defined as the question being clearly and unambiguously stated, and the answer being unique.\\

\#\#\# Examples \#\#\#\\
\textcolor{blue}{\{Few-shot Examples\}}\\

\#\#\# QA-pair \#\#\# \\
\textcolor{blue}{\{Input QA-pair\}}\\

\#\#\# Your Judgment \#\#\#

\end{tcolorbox}

\begin{tcolorbox}[title=Instruction of Graduate-level Filtering,width=\linewidth, breakable]

\#\#\# Instruction \#\#\#\\
You are a domain expert proficient in various subjects such as Math, Physics, Biology, Humanities, Computer Science, Engineering, and Chemistry. You will be given a question and its corresponding answer. Please identify whether the given QA-pair contains "Expert-level Knowledge." Only output 1 for yes and 0 for no. 

\#\#\# Definition \#\#\#\\
"Expert-level Knowledge" is defined as core theories, cutting-edge research, and complex applications studied during advanced undergraduate and graduate levels, typically mastered only by experts or senior researchers in the field.\\

\#\#\# Examples \#\#\#\\
\textcolor{blue}{\{Few-shot Examples\}}\\

\#\#\# QA Pair \#\#\# \\
\textcolor{blue}{\{Input QA-pair\}}\\

\#\#\# Your Judgment \#\#\#

\end{tcolorbox}

%% file: prompts/reward.tex
\begin{tcolorbox}[title=Instruction of Reward Model (official Instruction from HLE),width=\linewidth, breakable]
Judge whether the following [response] to [question] is correct or not based on the precise and unambiguous [correct\_answer] below.\\

[question]: \textcolor{blue}{\{question}\}\\

[response]: \textcolor{blue}{\{response}\}\\

Your judgement must be in the format and criteria specified below:\\

extracted\_final\_answer: The final exact answer extracted from the [response]. Put the extracted answer as 'None' if there is no exact, final answer to extract from the response.\\

[correct\_answer]: \textcolor{blue}{\{correct\_answer}\}\\

reasoning: Explain why the extracted\_final\_answer is correct or incorrect based on [correct\_answer], focusing only on if there are meaningful differences between [correct\_answer] and the extracted\_final\_answer. Do not comment on any background to the problem, do not attempt to solve the problem, do not argue for any answer different than [correct\_answer], focus only on whether the answers match.\\

correct: Answer 'yes' if extracted\_final\_answer matches the [correct\_answer] given above, or is within a small margin of error for numerical problems. Answer 'no' otherwise, i.e. if there if there is any inconsistency, ambiguity, non-equivalency, or if the extracted answer is incorrect.\\

confidence: The extracted confidence score between 0|\%| and 100|\%| from [response]. Put 100 if there is no confidence score available.\\

\end{tcolorbox}